\documentclass[conference,compsoc]{IEEEtran}
\ifCLASSOPTIONcompsoc
  \usepackage[nocompress]{cite}
\else
  \usepackage{cite}
\fi
\usepackage{booktabs}
\usepackage{multirow}
\usepackage{tabularray}
\usepackage{subfig}
\usepackage{algorithm}

\usepackage{amsmath}
\usepackage{amssymb}
\usepackage{pifont}
\usepackage{xcolor}
\usepackage[hidelinks]{hyperref}

\ifCLASSINFOpdf
\else
\fi

\usepackage{algpseudocode}

\usepackage{graphicx}
\usepackage{booktabs}
\usepackage{amsmath}
\usepackage{bbm}
\usepackage{algpseudocode}
\usepackage{multirow}
\usepackage{xspace}
\usepackage{subfig}
\usepackage{amssymb, color, diagbox} 
\usepackage{cite}
\usepackage{xcolor}
\usepackage[hidelinks]{hyperref}
\usepackage{mathrsfs}
\usepackage{bm}
\newcommand{\system}{ControlLoc\xspace}

\newenvironment{customitemize}{%
  \begin{list}{$\bullet$}{%
    \setlength{\itemsep}{2pt}%
    \setlength{\parsep}{0pt}%
    \setlength{\topsep}{0.5pt}%
    \setlength{\partopsep}{0pt}%
    \setlength{\leftmargin}{2em}
    \setlength{\labelwidth}{1em}%
    \setlength{\labelsep}{0.5em}%
  }%
}{%
  \end{list}%
}

\makeatother

\usepackage{mdframed}
\usepackage{tcolorbox,colortbl}

\newtcolorbox{mtbox}[1]{
  left=0.5mm,
  right=0.5mm,
  top=0.5mm,
  bottom=0.5mm,
  colframe=black, 
  colback=white, 
  boxrule=0.5pt,
  title={#1},
  fonttitle=\bfseries,
  coltitle=black,
  attach title to upper={},
  width=\linewidth,
  sharp corners
}

\begin{document}
%
\title{ControlLoc: Physical-World Hijacking Attack on Visual Perception \\ in Autonomous Driving}

\author{{\rm Chen Ma}$^{\star, \dagger}$  \quad {\rm Ningfei Wang}$^{\star, \ddagger}$ \quad {\rm Zhengyu Zhao}$^{\dagger }$ \quad {\rm Qian Wang}$^{\mathparagraph}$ \quad {\rm Qi Alfred Chen}$^\ddagger$ \quad {\rm Chao Shen}$^{\dagger}$ \\
$^\dagger$Xi'an Jiaotong University \quad
$^\ddagger$University of California, Irvine \quad
$^{\mathparagraph}$Wuhan University
}


%


\maketitle
\begingroup\renewcommand\thefootnote{$^{\star}$ The first two authors have equal contributions to the paper}
\footnotetext{}
\endgroup



\begin{abstract}

Recent research in adversarial machine learning has focused on visual perception in Autonomous Driving (AD) and has shown that printed adversarial patches can attack object detectors. However, it is important to note that AD visual perception encompasses more than just object detection; it also includes Multiple Object Tracking (MOT). MOT enhances the robustness by compensating for object detection errors and requiring consistent object detection results across multiple frames before influencing tracking results and driving decisions. Thus, MOT makes attacks on object detection alone less effective. To attack such robust AD visual perception, a digital hijacking attack has been proposed to cause dangerous driving scenarios such as vehicle collisions or unnecessary emergency stops. However, this attack has limited effectiveness even in the digital domain, let alone in the physical world.

In this paper, we introduce a novel physical-world adversarial patch attack, \system, designed to exploit hijacking vulnerabilities in entire AD visual perception. \system utilizes a two-stage process: initially identifying the optimal location for the adversarial patch, and subsequently generating the patch that can modify the perceived location and shape of objects with the optimal location. Extensive evaluations demonstrate the superior performance of \system, achieving an impressive average attack success rate of around 98.1\% across various AD visual perceptions and datasets. This performance significantly exceeds that of the existing hijacking attack, achieving four times greater effectiveness. The effectiveness of \system is further validated in physical-world conditions, including real vehicle tests under different conditions such as outdoor light conditions with an average attack success rate of 77.5\%. AD system-level impact assessments are also included, such as vehicle collision, using industry-grade AD systems and production-grade AD simulators with an average vehicle collision rate and unnecessary emergency stop rate of 81.3\%. We hope that our findings and insights can inspire future research into this safety- and security-critical domain.

\end{abstract}
\section{Introduction}
\label{sec:intro}

Autonomous Driving (AD) vehicles, also known as self-driving cars, are increasingly becoming an integral part of our daily lives~\cite{ubereat, taxila, zooxnv}. Various companies~\cite{adranking}, such as Tesla, are at the forefront of developing AD technologies. To ensure security and safety, AD vehicles employ visual perception to detect environmental elements such as traffic signs, pedestrians, and other vehicles in real time. These visual perception systems predominantly involve Deep Neural Networks (DNNs)~\cite{redmon2018yolov3, apollo, kato2018autoware} such as real-time object detection, owing to the superior performance of DNNs. 

Given that failing to detect the objects can lead to violent crashes~\cite{teslatw, uberarizona}, AD visual perception (referred to as AD perception throughout this paper) in ensuring safety and security has prompted extensive research into exploring its vulnerabilities. For instance, previous studies have highlighted the potential for adversarial attacks, including the use of adversarial patch~\cite{wen2024opticloak,jia2022fooling,zhao2019seeing, Wang_2023_ICCV,  eykholt2018robust}, to fool object detection in AD perception. Such attacks cause the AD systems to ignore objects, posing significant safety risks.

However, it is essential to recognize that AD perception extends beyond object detection to include Multiple Object Tracking (MOT)~\cite{apollo, kato2018autoware, jia2020fooling, shen2022sok}. MOT plays a pivotal role in AD perception by enhancing robustness against object detection errors. It ensures that only objects detected with consistent and stable accuracy across multiple frames are considered in the tracking results and, consequently, the driving decisions. Specifically, MOT tracks detected objects, estimates their velocities, and generates movement trajectories, called trackers. The tracker management module adds a layer of robustness against detection inaccuracies by not hastily discarding unmatched trackers or instantly creating new ones for newly detected objects. This multi-frame consistency requirement presents a significant challenge to attacks that solely target object detection. Thus, for an adversarial attack to significantly impact the AD perception pipeline, it must achieve at least a 98\% success rate across 60 consecutive frames~\cite{jia2020fooling}, which is infeasible for previous attacks on object detection~\cite{wen2024opticloak,zhao2019seeing,jia2022fooling, Wang_2023_ICCV}.

Therefore, a digital adversarial hijacking attack~\cite{jia2020fooling} to fool the entire AD perception has been proposed with adversarial patches as the attack vector. This hijacking attack necessitates precise control over the position and shape of the object, which is more challenging compared to prior attacks~\cite{cao2021invisible, sato2021dirty, zhu2023tpatch, jia2022fooling} that focus solely on manipulating an object confidence or classification scores. This attack is also powerful since it can achieve a persistent attack effect lasting for dozens of frames with just a few frames of successful attacks. Such a lasting impact is particularly valuable, as existing attacks on object detection alone require consistent success to achieve similar significant attack impacts. Despite this potential, prior attack~\cite{jia2020fooling} has shown limited effectiveness even in the digital space not to mention in the physical world since attacks generally face greater challenges in the physical world compared to the digital domain~\cite{cao2021invisible, sato2021dirty, zhu2023tpatch}. Such limited effectiveness of the prior hijacking attack is shown in~\S\ref{sec:baseline} across various AD perception modules and datasets. This limitation arises from their fundamentally ineffective methodology that fails to simultaneously optimize the necessary loss functions for conducting a successful attack. The detailed analysis and experiment results of the ineffectiveness of this adversarial hijacking attack~\cite{jia2020fooling} are demonstrated in~\S\ref{sec:tracker-hijacking-attack} and~\S\ref{sec:baseline}.

In this paper, we propose the first physical-world adversarial hijacking attack named \system on the \textit{entire AD perception} using adversarial patches, which can effectively control both the location and shape of objects. 
Our \system focuses on two attack goals: the move-in and the move-out attack. The objective of these attacks is to induce false perception in AD systems, potentially leading to dangerous driving scenarios such as vehicle collisions or unnecessary emergency stops, thereby affecting safety and traffic flow.

\system adopts a strategic two-stage approach. In the initial stage, we focus on finding the most effective location for placing the adversarial patch to facilitate successful hijacking attacks, formulating this task as an optimization problem. Subsequently, the second stage is to generate the adversarial patch, guided by the optimal locations identified in the preceding phase. This step involves erasing the target object's bounding box (BBOX) from the detection outputs, with a fabricated BBOX of a similar shape in a direction specified by the attacker based on the attack goals and scenarios. This process is designed to simulate movement in a chosen direction, deceiving the AD perception. 

To achieve the BBOX erasure and fabrication in the second stage, it is imperative first to determine the target fabricated BBOX, where we leverage the iterative process detailed in~\S\ref{sec:target_box}. We then design a BBOX filter detailed in~\S\ref{sec:box_filter} to eliminate extraneous proposal BBOXes. Moreover, we propose two loss functions, introduced in~\S\ref{sec:loss_design}, aimed at generating the adversarial patch to achieve the attack goal: a score loss, which controls the appearing or disappearing of the bounding boxes, and a regression loss, which is for shape and positioning of fabricated BBOX. Given the inherent challenges arising from the interdependence of these loss functions, we propose a novel optimization strategy, detailed in~\S\ref{sec:loss_design}, which demonstrates superior performance compared to existing optimization methods in prior research~\cite{jia2020fooling, jia2022fooling, Wang_2023_ICCV, zhao2019seeing}. Due to these designs, \system can significantly outperform the existing hijacking attack on AD perception~\cite{jia2020fooling}. Furthermore, \system uses the Expectation over Transformation (EoT) to enhance the robustness and effectiveness, especially in the real world.

Our evaluation results demonstrate that \system achieves outstanding performance across all different tested AD perceptions, including the combinations between four object detectors and four MOT algorithms for the two attack goals mentioned above. It is important to note that we include the AD perception adapted in open-source industry-grade full-stack AD systems~\cite{apollo, kato2018autoware}. On two driving datasets, \system realizes an impressive average attack success rate of 98.1\%. Furthermore, when compared with a baseline method~\cite{jia2020fooling}, our attack success rate is quadruple that of the baseline, underscoring its superior effectiveness. Additionally, our newly proposed optimization method in this problem domain surpasses the previous method by demonstrating the trend of different loss function values. 

To understand the attack effectiveness in the real world, we further evaluate \system with a real vehicle, where we put the generated adversarial patch on the rear of the car (and the location is specified by our patch location preselection in the first stage). The results show a 77.5\% average attack success rate across different outdoor backgrounds, light conditions, hijacking directions, and attack goals, while the AD perception can behave normally in all benign cases. To assess how \system affects the AD behavior, such as collisions or unnecessary emergency stops, we conduct tests using Baidu Apollo~\cite{apollo}, an industry-grade full-stack AD system with the LGSVL simulator~\cite{rong2020lgsvl}, a production AD simulator with an average effectiveness of 81.3\%. We also evaluate various existing directly adaptable DNN-level defense strategies and discuss future defense directions.

To sum up, we make the following contributions:
\begin{customitemize}
\item We propose the first practical hijacking attack on AD perception using physical-world adversarial patches as attack vectors, focusing on altering the location and shape of objects. This attack can potentially cause vehicle collisions or unnecessary emergency stops.
\item We introduce a novel attack framework, \system, to generate physical-world adversarial patches. This includes patch location preselection, BBOX filters, loss function designs, and a novel optimization method.
\item We evaluate \system on multiple AD perception systems including industry-grade ones. \system is effective in the real world with a real vehicle across different backgrounds, outdoor light conditions, hijacking directions, and scenarios. It causes AD system-level effects like collisions in a production AD simulator. 
\end{customitemize}

\section{Background and Related Work}
\label{sec:background}
In this section, we introduce visual perception in autonomous driving (AD) and attacks on AD visual perception.

\subsection{Visual Perception in Autonomous Driving (AD)}
\label{sec:perception}
\begin{figure}[!t]
\centering
\includegraphics[width=\linewidth]
{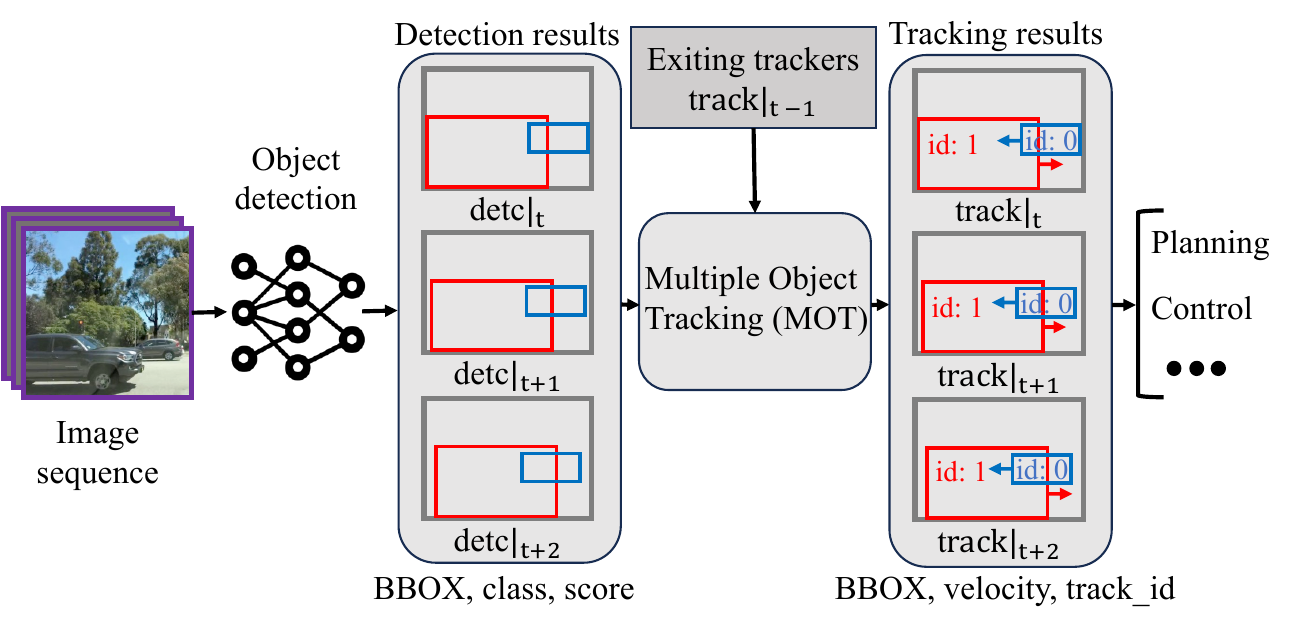}
\vspace{-0.6cm}
\caption{AD system pipeline: The camera captures images for object detection and multiple object tracking (MOT). Results are sent to the planning and then control modules.}
\label{fig:perception_overview}
\vspace{-0.4cm}
\end{figure}

Visual perception in AD (referred to as AD perception throughout this paper) critically depends on object detection and multiple object tracking (MOT) to accurately recognize and classify surrounding entities, such as cars. As depicted in Fig.~\ref{fig:perception_overview}, the process initiates with a series of images. The AD perception algorithm employs an object detector~\cite{zou2023object} to generate a bounding box (BBOX) and classify the object. Subsequently, the results from object detection, combined with existing tracking data, are input into the MOT~\cite{challa2011fundamentals}. This is tasked with updating the tracking information, such as the BBOX, object velocity, and track identification (track id). Finally, this data is relayed to other downstream modules in AD, such as the planning module~\cite{yurtsever2020survey}, which facilitates decision-making processes. Since only the detection results with sufficient consistency and stability across multiple frames can be included in the tracking results, the MOT module can improve the robustness of AD perception.

{\bf Object Detection}
Object detection plays a pivotal role in AD perception, predominantly utilizing Deep Neural Networks (DNN) to identify or categorize various road objects~\cite{carranza2020performance}. State-of-the-art DNN-based object detectors are fundamentally divided into two main categories: one-stage and two-stage detectors~\cite{zou2023object}. One-stage detectors, such as YOLO~\cite{Jocher2022yolov5, redmon2018yolov3, redmon2017yolo9000}, are renowned for their rapid detection speeds, making them highly suitable for real-time applications such as AD systems. In contrast, two-stage detectors, such as Faster R-CNN\cite{ren2015faster}, are celebrated for their accuracy in detection. Given the real-time requirement of AD systems, industry-grade full-stack AD systems, such as Baidu Apollo~\cite{apollo}, predominantly employ one-stage detectors. Furthermore, object detection can be categorized into anchor-based and anchor-free approaches~\cite{zhang2020bridging}. Anchor-based detection methods leverage a large number of preset anchors, then predict the category and refine the coordinates of these anchors, and finally output these refined anchors as detection results. Conversely, anchor-free detection~\cite{tian2020fcos}, directly predicts the bounding boxes of objects, offering a more generalizable solution. This paper explores both anchor-based and anchor-free object detection methods.

{\bf Multiple Object Tracking (MOT)}
The current state-of-the-art MOT can be broadly classified into two main approaches~\cite{luo2021multiple, dhar2020object}: detection-based tracking, also known as tracking-by-detection, and detection-free tracking. The former method employs object detectors to identify objects, which are then used as inputs for MOT, while the latter relies on manually cropped or marked objects as inputs~\cite{luo2021multiple}. Tracking-by-detection has emerged as the predominant technique in MOT, particularly within the context of AD~\cite{luo2021multiple, dhar2020object, zhang2022bytetrack}. This predominance is attributed to the inherent unpredictability of the number and locations of objects, coupled with the expectation that objects can periodically enter and exit the camera field of view~\cite{dhar2020object}. These conditions render tracking-by-detection algorithms especially well-suited for integration into AD systems~\cite{dhar2020object}. In this paper, we concentrate on the tracking-by-detection paradigm. As illustrated in Fig.~\ref{fig:perception_overview}, this methodology involves associating the results of object detection at time $t$ with existing trackers from the previous time step ($track|{t-1}$) and forecasting the current state of the trackers at time $t$ ($track|{t}$), which includes the velocity and location of every tracked object. To mitigate the impact of false positives and missed detection by the object detectors, MOT modules typically initiate a tracker for an object only after it has been consistently detected across $H$ frames. Similarly, a tracker is removed only after the object has not been detected for $R$ consecutive frames~\cite{zhu2018distractor, apollo, kato2018autoware, jia2020fooling}. Consequently, merely compromising the object detection component may not sufficiently disrupt the AD perception~\cite{shen2022sok, jia2020fooling}. Therefore, this paper introduces a novel and practical physical-world adversarial hijacking attack strategy targeting the entire AD perception including both object detection and MOT.

\begin{figure*}[!t]
\centering
\includegraphics[width=\linewidth]
{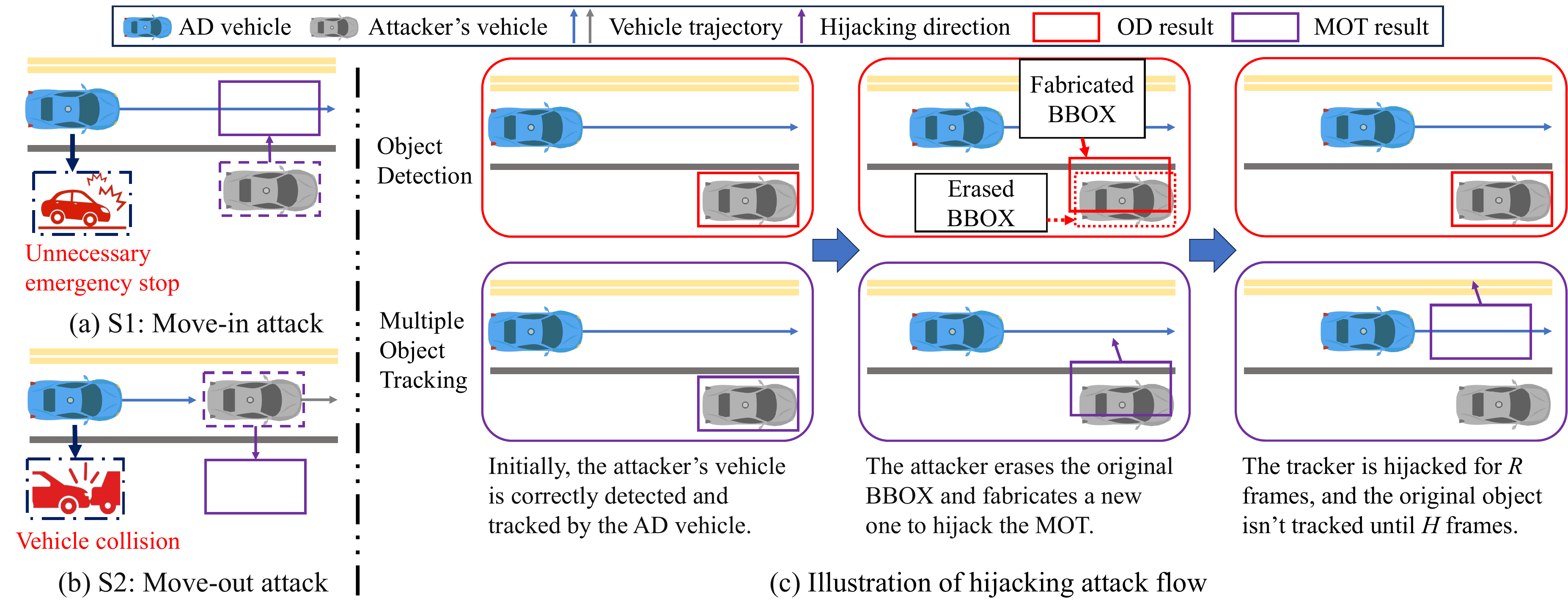}
\vspace{-0.6cm}
\caption{Attack goals: (a) move-in attack and (b) move-out attack. (c) shows hijacking attack flow.}
\label{fig:ill}
\vspace{-0.4cm}
\end{figure*}

\subsection{Attacks on AD Visual Perception}
\label{sec:literature}

{\bf Attacks on Object Detection} 
\label{sec:ae}
Recent studies have highlighted the vulnerability of DNN models to adversarial examples or attacks, a vulnerability that has been extensively explored~\cite{goodfellow2014explaining, carlini2017towards, papernot2016limitations, zhang2020interpretable, xie2017adversarial}. Further investigations have extended these findings to adversarial attack in the physical world~\cite{wen2024opticloak,jia2022fooling,eykholt2018robust, zhao2019seeing, Wang_2023_ICCV, lovisotto2021slap, cao2021invisible, sato2021dirty, xu2020adversarial}. Specifically, within the context of AD, prior research has successfully executed physical-world adversarial attacks targeting visual object detection alone~\cite{ wen2024opticloak,jia2022fooling, Wang_2023_ICCV, zhao2019seeing}. However, the entire AD perception framework encompasses both object detection and MOT. Given the nature of MOT, for an attack targeting only object detection to be effective, it must achieve at least a 98\% success rate across 60 consecutive frames—a highly challenging task that for existing object detection attacks to meet~\cite{wen2024opticloak,jia2020fooling, muller2022physical}. Thereby, this paper proposes a novel and practical method for the physical-world adversarial hijacking attack aiming at the entire AD perception to enhance the attack's effectiveness.

{\bf Attacks on Object Tracking}
\label{sec:MOT_attack}
Various attacks targeting object tracking have been proposed, spanning both the digital~\cite{chen2021unified, yan2020hijacking} and physical domains~\cite{muller2022physical, ding2021towards}. Among them, AttrackZone~\cite{muller2022physical} represents a notable physical domain attack against siamese-based tracking~\cite{li2018high, li2019siamrpn++}, which is a single object tracking (SOT)~\cite{zhang2021recent}, employing a projector to introduce adversarial perturbations. However, contemporary AD systems employ MOT rather than SOT~\cite{apollo, kato2018autoware, chen2021online, math9060660, shen2022sok} due to the requirement to identify and track multiple objects simultaneously~\cite{math9060660}. In addition, AttrackZone operates under a white-box threat model with siamese-based trackers. Although AttrackZone demonstrates a notable attack success rate in transferability between different siamese-based SOT DNN models, adapting this methodology to MOT used in AD perception is challenging~\cite{apollo, kato2018autoware, chen2021online, math9060660, shen2022sok} since gradient information is generally not available. Given the limited application of SOT in AD contexts, the end-to-end impact of AttrackZone on AD vehicles is uncertain, casting doubts on its practical value. In contrast, in this paper, we introduce a novel attack against the realistic entire AD perception, i.e., object detection plus MOT, leveraging an adversarial patch effective in different light conditions and backgrounds. Our attack adopts a black-box threat model for the object tracking algorithm detailed in~\S\ref{sec:threat-model-attack-goal}, offering a more realistic approach compared to AttrackZone. Our attack is notable for a certain level of generality and practicality, alongside a novel attack optimization method.

{\bf Attacks on the Entire AD Perception} 
Attacks targeting the entire AD perception, such as on availability~\cite{ma2023slowtrack} and integrity~\cite{jia2020fooling}, have been documented in the literature. Notably, Jia et al.~\cite{jia2020fooling} introduce a digital adversarial hijacking attack for the entire AD perception, encompassing object detection and MOT. Despite the innovative approach, the effectiveness of their attack is fundamentally limited as shown in~\S\ref{sec:baseline} even in the digital space not to mention in the physical domain. In contrast, our work presents a physical-world hijacking attack that not only breaks the AD perception but also enhances effectiveness compared to the prior attack.

\section{Attack Goal and Threat Model}
\label{sec:threat-model-attack-goal}

{\bf Attack Goal.} In this paper, we primarily focus on attack goals with significant safety implications for AD, such as vehicle collisions or unnecessary emergency stops~\cite{wan2022too}. We specifically explore physical-world attack vectors within the AD landscape, employing the adversarial patch due to its high practicality and realism~\cite{zhao2019seeing, Wang_2023_ICCV, eykholt2018robust, zhu2023tpatch}. Our research outlines two main hijacking attack goals: the move-in attack and the move-out attack, shown in Fig.~\ref{fig:ill} (a) and (b), respectively. The move-in attack is designed to deceive the victim AD vehicle into an unnecessary emergency stop by inducing a false perception of an object on its current trajectory. On the other hand, the move-out attack manipulates the AD system to overlook actual obstacles by altering the perceived location of these obstacles to the roadside, thereby leading the vehicle into a collision. These tactics aim to demonstrate the potential for adversarial interventions to disrupt the safety and operational integrity of AD systems.

{\bf Threat Model.} To achieve the attack goals outlined above, this paper delves into a white-box threat model for object detection but a black-box threat model for multiple object tracking (MOT) within AD perception. This threat model assumes that the attacker possesses detailed knowledge of the target object detection, including its architecture and parameters, a promising threat model that aligns with the ones in the existing literature on adversarial vulnerabilities of AD perception~\cite{jia2020fooling, jia2022fooling, cao2021invisible, sato2021dirty, muller2022physical, zhao2019seeing, eykholt2018robust}. The black-box assumption on MOT is justified by the generally similar designs across tracking-by-detection algorithms used in AD systems~\cite{zhang2022bytetrack, apollo, kato2018autoware}. 
To effectively attack the AD perception, we leverage the adversarial patch, which is physically realizable~\cite{zhao2019seeing, dipalma2021security, Wang_2023_ICCV, jia2022fooling, eykholt2018robust}. Examples include physically printing the patch~\cite{jia2022fooling, zhao2019seeing, Wang_2023_ICCV, eykholt2018robust} or displaying the patch on a monitor attached to the back of a vehicle~\cite{man2023person}. Such adversarial patch position and size can also be controlled by the attacker.

\section{\system Attack Methodology}
\label{sec:tracker-hijacking-attack}

In this section, we present the first physical-world hijacking attack: \system on the entire AD perception including object detection and MOT.

\begin{figure*}[!t]
\centering
\includegraphics[width=0.95\linewidth]{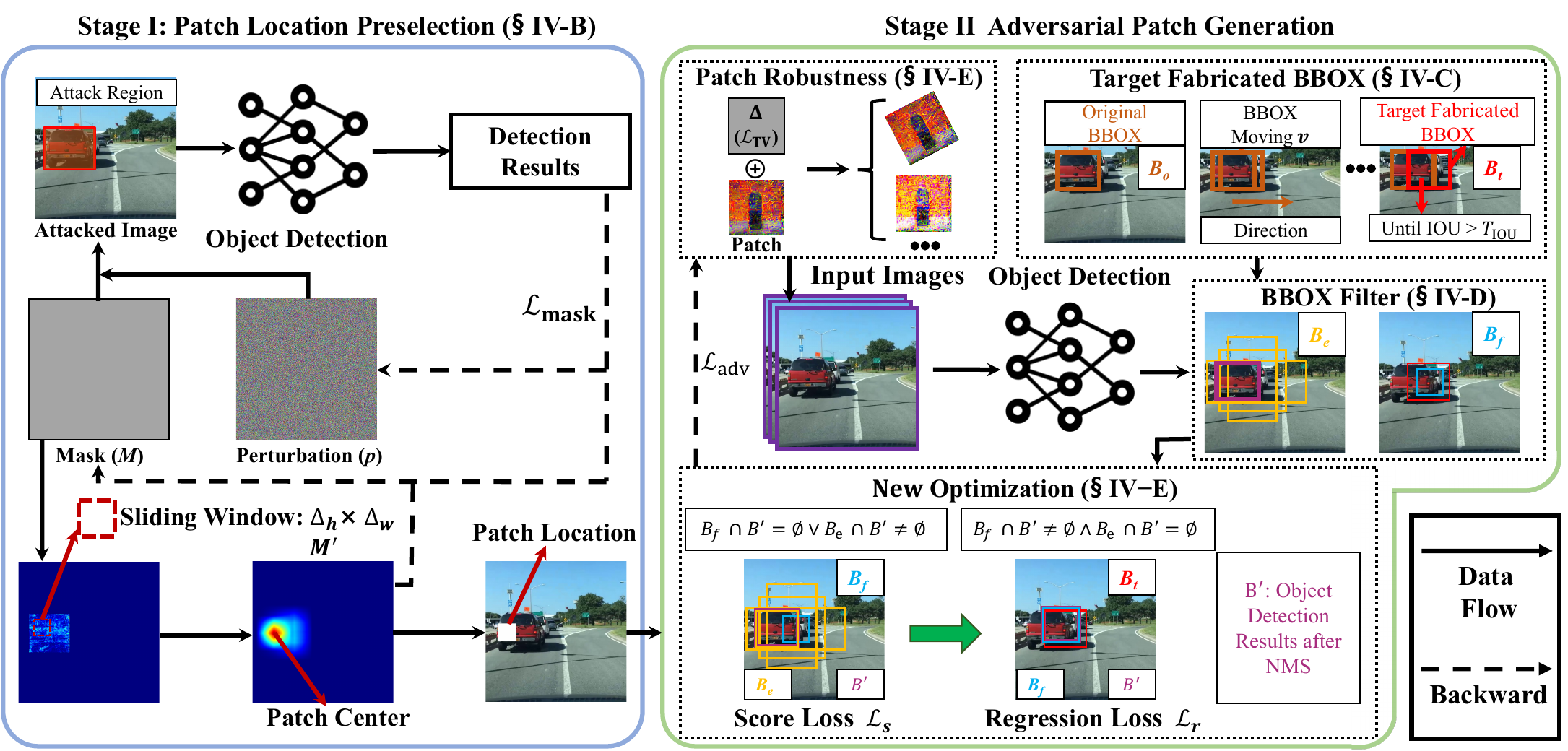}
\caption{Overview of our \system, a two-stage hijacking attack via adversarial patches.}
\label{fig:fig_overview}
\vspace{-0.4cm}
\end{figure*}

\subsection{Attack Design Overview}
\label{sec:attack_design_overview}
We provide a detailed overview of our \system. This hijacking attack flow is illustrated in Fig.~\ref{fig:ill} (c). As depicted, the process begins with the AD perception system correctly detecting and tracking the object. When the vehicle enters the effective attack range, \system removes the bounding box (BBOX) of the target object from the detection results and fabricates a similar-shaped BBOX, which is slightly shifted with an attacker-desired direction. This fabricated BBOX is then associated with the original tracker of the target object, effectively hijacking the tracker. Although the tracker hijacking typically lasts for only a few frames, its adversarial effects can persist longer, depending on the design of the MOT, particularly the common $H$ and $R$ parameters shown in Fig.~\ref{fig:ill} (c) and introduced in~\S\ref{sec:background}.
To achieve the above attack strategy, we propose a dual-stage attack method, of which overview is shown in Fig.~\ref{fig:fig_overview}.

{\bf Stage I:} This stage shown in Fig.~\ref{fig:fig_overview} is an optimization-based approach to preselect the patch location. The details of this part will be introduced in~\S\ref{sec:patch_location}). This strategy leverages masks and adversarial perturbations to identify areas that are most conducive to successful attack execution. These areas are then further refined based on potential patch placement locations, such as the rear of the vehicle. Subsequently, a sliding window is utilized to precisely obtain the optimal location. Stage I can be a pre-processing step to enhance the efficiency and effectiveness of attack generation.

{\bf Stage II:} This stage shown in Fig.~\ref{fig:fig_overview} can be divided into several distinct steps, outlined below, focusing on generating a physical-world adversarial patch for hijacking attacks.

\textit{{Step 1: Finding Target Fabricated Bounding Box.}} In Fig.~\ref{fig:fig_overview}, an iterative process is employed to find the target fabricated BBOX based on the Intersection over Union (IOU) value between the candidate and the original BBOX. The key insight is that the fabricated BBOX should closely match the original BBOX, but with a shift as large as possible towards the attack direction. The details are outlined in~\S\ref{sec:target_box}.

\textit{{Step 2: Bounding Box Filter.}} In DNN-based object detection, many proposed BBOXes are irrelevant for attack generation, often identifying background elements or unrelated objects. To ensure the effective generation of the patch, it is crucial to filter the relevant BBOXes. This BBOX filter process is conducted based on the understanding of the object detection designs and is elaborated upon in~\S\ref{sec:box_filter}.

\textit{{Step 3: Loss Function Design and Optimization Method.}} This step introduces novel loss functions and a new optimization method detailed in~\S\ref{sec:loss_design}. The designed loss function includes score loss and regression loss to create or remove BBOXes. We propose a new optimization strategy that markedly enhances the effectiveness of the traditional standard Lagrangian relaxation method. To bolster attack robustness, we integrate Expectation over Transformation (EoT), drawing upon prior research~\cite{Wang_2023_ICCV, zhao2019seeing}.

\subsection{Patch Location Preselection}
\label{sec:patch_location}
To effectively generate our attack, it is crucial to strategically position a patch in the most vulnerable area near the vehicle. We formulate this problem as an optimization problem to find the ideal region for patch placement~\cite{cheng2023fusion}. The detailed process is illustrated in Fig.~\ref{fig:fig_overview}. The objective function, denoted as $\mathcal{L}_\mathrm{mask}$, is formulated as follows:
\begin{align}
         ~&\arg\min\limits_{p, m} \mathcal{L}_\mathrm{adv} (x') +\alpha\cdot\mathcal{L}_{M^{\prime}} (m,\Delta_h,\Delta_w) \label{eq:mask}\\        \mathbf{where}~&\mathcal{L}_{M^{\prime}}=\|\max(M^\prime)-  \frac{1}{hw}\sum\limits_{j=1}^h\sum\limits_{i=1}^{w}M^\prime[i,j]\|_1 \\
        ~&M^\prime[i, j] = \sum\limits_{z_1 = 0}^{\Delta_h - 1}\sum\limits_{z_2 = 0}^{\Delta_w - 1}M[i + z_1, j + z_2]W[z_1,z_2]\label{eq:mprime}\\
        ~&M[i, j] = \frac{1}{2}\times\tanh(\gamma\cdot m[\lfloor \frac{i}{s}\rfloor, \lfloor \frac{j}{s} \rfloor])+\frac{1}{2} \label{eq:mask_m}\\
        ~&\hspace{20pt}x' = x\odot(1-M) + p \odot M\label{eq:x_prime}
\end{align} 
Equation~\eqref{eq:mask} is to identify the most vulnerable region leveraging the mask denoted as $M$, which controls the strength of the perturbation $p$. The final patch location aims to contain as many pixel points with high values of $M$ as possible, to cover the most vulnerable areas. When using this method, two main concerns must be addressed. First, the values of $M$ need to be kept as close to 0 or 1 as possible to reflect the binary decision of either applying or not applying the patch at each pixel point. Second, it is important to keep that the high-value pixel points of 
$M$ are clustered closely together, as the patch needs to form a contiguous block.

To address the first concern, $M$ is computed by unconstrained mask parameters $m$, as shown in Equation~\eqref{eq:mask_m}. The transformation using the $\tanh$ function in Equation~\eqref{eq:mask_m} constrains the mask $M$ within [0, 1] range. Tuning the hyperparameter $\gamma$ drives mask values closer to 0 or 1 and modulates the convergence speed of the process. The hyperparameter $s$ modulates the mask granularity. The variable $p$ signifies the perturbations applied to the original image $x$ through Equation~\eqref{eq:x_prime} to obtain the input $x'$ for $\mathcal{L}_\mathrm{adv}$. The details of $\mathcal{L}_\mathrm{adv}$ will be introduced in~\S\ref{sec:loss_design}. The variables $h$ and $w$ represent the height and width of the image $x$.

To address the second concern, upon generating a sensitivity mask indicative of the perturbation mask $M$, a sliding window $W$ of the same size $(\Delta_h,\Delta_w)$ as the patch is applied to process this mask. The calculated averaged values within the window are referred to as $M^{\prime}$ shown in Equation~\eqref{eq:mprime}, which scores each potential location by averaging the values within the window. Furthermore, leveraging the mask $M^{\prime}$, we formulate a novel loss function $\mathcal{L}_{M^{\prime}}$, which plays a pivotal role in determining the unique and most effective patch location. Specifically, by minimizing $\mathcal{L}_{M^{\prime}}$, we can encourage $M$ to cluster within a uniquely rectangular box of dimensions $(\Delta_h,\Delta_w)$. The clustering effect is super important for the effectiveness of an adversarial patch, as the patch must form a contiguous block. Therefore, \system focuses on finding the optimal placement for a tightly grouped continuous block rather than scattered discrete points. $\alpha$ is a hyperparameter.

Moreover, recognizing physical constraints on the capabilities of the attackers, only designated areas, such as the rear of the vehicle, are considered viable for patch placement. Thereby, we limit the mask $M$ to these regions. Notably, selecting the patch location can precede attack generation steps, serving as a potential and effective pre-processing step. Furthermore, our patch selection method incurs negligible computational overhead, as we only required 20 iterations in our experiments to determine the optimal location shown in our evaluation~\S\ref{sec:baseline}.

\subsection{Finding Target Fabricated Bounding Box}
\label{sec:target_box}

The core idea behind finding a target fabricated BBOX is to create a scenario where, when our attack has ended, the tracking system loses track of the original object. This is achieved by manipulating the BBOX of the target object to maximize its deviation from the benign, within its original data association range,
directing towards a directional vector $\Vec{v}$ determined by the attack goal. Unlike previous research~\cite{jia2020fooling}, which seeks the optimal BBOX location based on the specific tracking algorithm, we employ a tracking-agnostic strategy based on the observation that the adversarially modified BBOX does not require precise alignment with the adversarial patch's physical location.

To achieve that, the key insight of this approach is that the fabricated BBOX should match the original BBOX, but with a shift as large as possible towards the direction $\Vec{v}$. Thus, the fabricated BBOX must overlap the benign BBOX with an IOU above a predefined threshold $T_\mathrm{IOU}$, while also being slightly shifted towards the original BBOX position. It's noteworthy that this IOU threshold generally remains consistent across different MOT~\cite{du2023strongsort, zhang2022bytetrack, jia2020fooling, apollo, kato2018autoware}, enabling the application of a general threshold that facilitates a black-box attack model. This general property is critical, as it does not require detailed knowledge of the specific MOT algorithms in use. Our method for iteratively determining the target fabricated BBOX location is outlined in Algorithm~\ref{alg:find_pos} to find the desired deviation and illustrated in Fig.~\ref{fig:fig_overview}.

\begin{algorithm}[!t]
  \caption{Find target fabricated BBOX location} 
  \label{alg:find_pos}
  \begin{algorithmic}[1]
    \Require
      $B_o$: Original object BBOX;
      $\Vec{v}$: Attacker desired directional vector;
      $T_\mathrm{IOU}$: IOU threshold for data association between trackers and detection results.
    \Ensure
      $B_t$: Target fabricated BBOX location.
    \State $k \leftarrow 1$
    \State $B_t \leftarrow B_o$
    \While{$\mathrm{IOU}(B_t, B_o) > T_\mathrm{IOU}$}
        \State $B_t\leftarrow B_o + \Vec{v} \cdot k$
        \State $k=k+1$
    \EndWhile
    
    \State $B_t\leftarrow B_o + \Vec{v} \cdot (k-1)$

    \State \Return $B_t$
  \end{algorithmic}
\end{algorithm}

\subsection{Bounding Box Filter}
\label{sec:box_filter}

In DNN-based object detection, most proposed BBOXes do not contribute to patch generation, as they frequently identify irrelevant objects or background elements. To generate the patch effectively, it requires the selection of appropriate BBOXes for fabrication or erasure. This selection hinges on the understanding of object detection. Our approach is adaptable to both anchor-based and anchor-free detection models as introduced in~\S\ref{sec:perception}.
%

The mainstream object detectors, including one-stage detectors such as the YOLO series, or two-stage detectors such as the RCNN series, introduced in~\S\ref{sec:background}, can adopt grid-based designs~\cite{redmon2018yolov3, redmon2017yolo9000, Jocher2022yolov5, lu2019grid, he2017mask, ren2016faster}. Grid-based detectors separate the input image into fixed-size grids, with each cell responsible for predicting BBOXes for objects within its vicinity. To ascertain the location of these BBOXes, an offset is calculated from the top-left corner of each cell. A detailed illustration and example for this process are provided in Appendix~\ref{sec:yolo5}, which precisely obtains the BBOX location.

By leveraging the intrinsic property of grid-based detectors above, we introduce the Center bounding box filter (C-BBOX), an effective method for filtering BBOX adaptable for both anchor-based and anchor-free object detection detailed in~\S\ref{sec:background}. The details of the C-BBOX process are in Algorithm~\ref{alg::C-BBOX}. C-BBOX first calculates the scaling ratio $\mathrm{scale}$ between the input image size and the feature map size, i.e., the size of each grid cell. Then the C-BBOX extracts the grid cell corresponding to $B_t$ (\S\ref{sec:target_box}) based on $\mathrm{scale}$. 

C-BBOX is compatible with anchor-based and anchor-free models. For anchor-based detectors, where each grid corresponds to multiple anchors, C-BBOX extracts the BBOX having the largest IOU with $B_t$ as $B_f$ in Algorithm~\ref{alg::C-BBOX} ($\mathrm{top}(A)$ is to obtain the index of maximum value in vector $A$). For anchor-free detectors, where each grid has a unique anchor, C-BBOX applies a corrective vector in the hijacking direction to accurately filter the BBOXes since such detectors allow for greater flexibility in BBOX placement.

Moreover, C-BBOX assists in pinpointing BBOXes for erasure in anchor-based models by identifying the cell corresponding to the original BBOX, thereby enabling the precise removal of undesired BBOXes. For anchor-free detectors, we use the IOU BBOX filter, similar to previous research~\cite{jia2022fooling}, to identify BBOXes for erasure. This method initially eliminates predictions with confidence below the NMS threshold. Subsequently, it filters the BBOX by the IOU between each remaining proposal BBOX and $B_t$.

\begin{algorithm}[!t]
  \caption{C-BBOX} 
  \label{alg::C-BBOX}
  \begin{algorithmic}[1]
    \Require
      $B_t$: Target BBOX;
      $\mathrm{size}_{f}$: Feature map size;
      $\mathrm{size}_{x}$: Image size;
      $\Vec{v}$: Attack directional vector,
      $k_s$: Step size.
    \Ensure
      $B_f$: BBOX needed to be fabricated.
    \State $(c_x,c_y)\leftarrow \text{center point of }B_t$
    \State $\mathrm{scale} = \mathrm{size}_{x} / \mathrm{size}_{f}$
    \State $\mathrm{id_{grid}}=\textbf{int}(c_x/\mathrm{scale},c_y/\mathrm{scale})$
    \State $\mathrm{grid}\leftarrow \text{grid cell corresponding to $\mathrm{id_{grid}}$}$
    \If  {$\mathrm{detector\ is\ anchor\_based}$}
    \State $\mathrm{anchors}\leftarrow \text{all anchors of $\mathrm{grid}$}$
    \State $\mathrm{index_{anchor}}=\textbf{{top}}(\mathrm{IOU}(B_t,\mathrm{anchors}))$;
    \State $B_f=\mathrm{anchors[index_{anchor}]}$
    \ElsIf {$\mathrm{detector\ is\ anchor\_free}$}
    \State $c_x=c_x+ k_s \cdot \frac{\Vec{v}}{|\Vec{v}|}$
    \State $\mathrm{id_{grid}}=\textbf{int}(c_x/\mathrm{scale},c_y/\mathrm{scale})$    
    \State $B_f\leftarrow \text{the anchor of $\mathrm{grid}$}$ corresponding to $\mathrm{id_{grid}}$
    \EndIf

    \State \Return $B_f$
  \end{algorithmic}
\end{algorithm}

For detectors not based on grid structures, bipartite matching~\cite{carion2020end}, is used to distinguish between BBOXes for fabrication and those for erasure. This approach ensures our method's applicability across various object detection designs. With the filter methods in Equation~\eqref{eq:eq2}, we extract the BBOXes needed to be fabricated $B_f$ and erased $B_e$.
\begin{equation}
\label{eq:eq2}
\centering
{B_f}, {B_e} = {F(O_\mathrm{bbox},B_t,B_o)}
\end{equation}
where $O_\mathrm{bbox}$ is all proposal BBOXes before NMS, $B_o$ is the original BBOX, and $F(\cdot)$ is the BBOX filter function.

\begin{algorithm}[!t]
  \caption{Generating Adversarial Patch} 
  \label{alg::conjugateGradient}
  \begin{algorithmic}[1]
    \Require
       $x$: Input image;
       $B_t$: Target BBOX;
       $B_o$: Original object;
       $D(\cdot)$: Object detector;
       $N$: Attack iterations;
       NMS($\cdot$): NMS function;
       $T_{conf}$: Score threshold.
    \Ensure
       $\Delta$: Adversarial patch.
    \State Initial $\Delta \leftarrow \Delta_0$ 
    \For{$n = 1$ to $N$}
        \State $O_\mathrm{bbox} = D(x+\Delta)$;
        \State $B_f, B_e = F(O_\mathrm{bbox}, B_t, B_o)$
        \State $B^\prime = \mathrm{NMS}(O_\mathrm{bbox})$
        \If{$\ B_f \cap B^\prime\neq\varnothing \ \text{and} \  B_e \cap B^\prime=\varnothing$}
            \State $\mathcal{L}_\mathrm{adv}=\mathcal{L}_r(B_f,B_t)$
            \Else
            \State $\mathcal{L}_\mathrm{adv}=\mathcal{L}_s(B_f,B_e,T_{conf})$
        \EndIf
        \State $\mathcal{L} = \mathcal{L}_\mathrm{adv} + \mu_2 \cdot\mathcal{L}_\mathrm{TV}$
        \State $\Delta = \mathrm{Adam}(\Delta,\mathcal{L})$
    \EndFor
    \State \Return $\Delta$
  \end{algorithmic}
\end{algorithm}

\subsection{Loss Design and Optimization Method}
\label{sec:loss_design}

As detailed in Algorithm~\ref{alg::conjugateGradient}, \system involves enhancing the confidence score of $B_f$ to ensure its preservation after NMS, while concurrently adjusting its dimensions and location to closely match $B_t$. Conversely, it is imperative to diminish the confidence scores of $B_e$ to preclude their inclusion in the detection outcomes. Similar to the existing adversarial patch attacks~\cite{zhao2019seeing, Wang_2023_ICCV}, we also formulate the adversarial patch generation as an optimization problem. The optimization of this attack poses a multiple-objective problem, requiring the simultaneous optimization of the score loss $\mathcal{L}_{s}$ for the extracted boxes as well as the shape and location loss, collectively referred to as regression loss $\mathcal{L}_{r}$. Specifically, for an input image $x$ and an object detection model $D(\cdot)$ that excludes NMS, the optimization task can be represented in Equation~\eqref{eq:eq3}, aiming to minimize $\Delta$ subject to the conditions that $B_f$ is encompassed within $B^\prime$ and $B_e$ is excluded from $B^\prime$, where $B^\prime$ is all BBOXes after NMS.
\begin{equation}
\label{eq:eq3}
\begin{array}{l}
\mathop{\arg\min}\limits_{\Delta} \mathcal{L}\{F[D(x, \Delta), B_t, B_o]\} \\ 
\ \ \ \ \ \text{s.t.} \ B_f \in B^\prime \ \text{and} \  B_e \cap B^\prime=\varnothing 
\end{array}
\end{equation}
where $\Delta$ is adversarial patch and $F$ is from Equation~\eqref{eq:eq2}.

{\bf Score Loss.} 
To effectively manipulate BBOXes in \system, adjusting their scores is essential. This adjustment aims to enhance the scores of newly generated BBOXes denoted as $\mathcal{L}_{f}$ while simultaneously reducing the scores of removed BBOXes denoted as $\mathcal{L}_{e}$. To accomplish this, we introduce a novel score loss, defined in Equation~\eqref{eq:score_loss}.
\begin{align}
    \mathcal{L}_{s} = &\underbrace{{\frac{1}{|{B}_{e}|}} \sum\limits_{{c}\in{B}_{e}}\mathop{\mathbbm{1}^c}\cdot \ c_{conf}^2}_{\mathcal{L}_{e}}  + \underbrace{\mu_{1}\cdot{\frac{1}{|{B}_{f}|}} \sum\limits_{{c}\in{B}_{f}}(1-c_{conf})^2}_{\mathcal{L}_{f}}\label{eq:score_loss} \\
    &c_{conf} = c_{obj} \cdot {\max}\{c_{class_i}\}, \ i \in [1, N_c]\label{eq:cconf}
\end{align}
where $N_c$ is the number of classes; the indicator function $\mathbbm{1}^{c}$ checks whether the score of a BBOX $c$ exceeds the score threshold $T_{conf}$; it is set to 1 if true, and 0 otherwise. This formulation aims to adjust scores, enhancing the detection of relevant objects while minimizing the impact of irrelevant ones. Hyperparameters $\mu_1$ is to balance $\mathcal{L}_{e}$ and $\mathcal{L}_{f}$.

{\bf Regression Loss.} To optimize the position and shape of the fabricated BBOXes $B_f$—aiming to effectively redirect the tracking from the target object—we introduce a regression loss function, as delineated in Equation~\eqref{eq:regr_loss}. 
\begin{equation}
\label{eq:regr_loss}
\begin{array}{l}
    \mathcal{L}_{r} = \overbrace{{\frac{1}{|{B}_{f}|}}\sum\limits_{{c}\in{B}_{f}}-\log(\mathrm{IOU}(c,B_t))}^{\mathcal{L}_\mathrm{IOU}} \\ 
\ \ \ + \ \beta \cdot \underbrace{{\frac{1}{|{B}_{f}|}}\sum\limits_{{c}\in{B}_{f}}(\mathrm{center}(c)-\mathrm{center}(B_t))^2}_{\mathcal{L}_\mathrm{center}}
\end{array}
\end{equation}
where the regression loss $\mathcal{L}_{r}$ comprises two components: $\mathcal{L}_{\mathrm{IOU}}$ and $\mathcal{L}_{\mathrm{center}}$. The IOU loss aims to reduce the discrepancy in the overlap between the fabricated BBOX $B_f$ and the target BBOX $B_t$, ensuring accurate coverage and alignment. Thus, the center loss, weighted by a factor $\beta$, seeks to minimize the distance between the centroids of $B_f$ and $B_t$ such that the tracker can be moved away.

{\bf Total variation Loss.} 
To make the generated adversarial patch smooth, and thus increase the effective range of the attack, the total variation loss in Equation~\eqref{eq:loss-tv} is used to reduce the color changes between the adjacent pixels. 
\begin{equation}
\mathcal{L}_\mathrm{TV}  = \sum_{i, j} \sqrt{\left|\Delta_{i+1, j}-\Delta_{i, j}\right|^{2}+\left|\Delta_{i, j+1}-\Delta_{i, j}\right|^{2}}
\label{eq:loss-tv}
\end{equation}

{\bf Optimization Method.} 
Simultaneously optimizing multiple loss functions, particularly $\mathcal{L}_{s}$ and $\mathcal{L}_{r}$, requires a sophisticated strategy. Existing literature~\cite{jia2020fooling} typically employs the standard Lagrangian relaxation method for this task. This approach involves aggregating the different loss functions into a single objective, each modulated by predetermined coefficients, followed by gradient descent to seek an optimal solution. Yet, this methodology requires adjustment of the coefficients corresponding to each loss function to achieve desired outcomes. The endeavor to manage multiple objectives within a unified optimization framework introduces complexity, leading to potential difficulties in balancing the influence of each loss function and thus low effectiveness. Therefore, achieving an optimal balance requires careful tuning to determine the most effective setting. 

In our case, this method is fundamentally ineffective. Notably, it does not perform well across various coefficient configurations, as detailed in~\S\ref{sec:baseline}. The inefficacy of simultaneously optimizing multiple loss functions, i.e., $\mathcal{L}_{s}$ and $\mathcal{L}_{r}$, is largely attributed to the negative coupling effects in gradients. Essentially optimizing $L_f$ in $L_s$ determines the location of $B_f$ at a coarse-grained level. Subsequent optimization of $L_r$ refines the location and shape of $B_f$. Thus, the appropriate sequence of optimization should initially focus on $L_f$ in $L_s$, to ensure that $B_f$ is correctly identified in the detection results after NMS. Then, the subsequent step involves adjusting the location and shape of $B_f$. However, employing the standard Lagrangian relaxation method to achieve dual optimization presents challenges. The interaction between $L_r$ and $L_f$ in $L_s$ often leads to a negative coupling effect in our problem space, where an excessive gradient on one side can restrict improvements in the other, hindering effective optimization.

Thus, we propose a new optimization method to address the limitations in the standard Lagrangian relaxation method for hijacking attack generation mentioned above:
\begin{equation}
\begin{aligned}
\label{eq:eq9}
~ &\mathop{\arg\min}\limits_{\Delta}\mathcal{L}_\mathrm{adv}+\mu_{2}\cdot\mathcal{L}_\mathrm{TV} \\
\mathbf{where \ } \mathcal{L}_\mathrm{adv} ~& = {\mathbbm{1}}[B_f \cap B^\prime\neq\varnothing  \text{ and } B_e \cap B^\prime=\varnothing] \cdot \mathcal{L}_r \\
& + {\mathbbm{1}}[B_f \cap B^\prime=\varnothing \text{ or } B_e \cap B^\prime\neq\varnothing] \cdot \mathcal{L}_s
\end{aligned}
\end{equation}

Our method optimizes either $\mathcal{L}_s$ or $\mathcal{L}_r$ based on the condition specified shown in Equation~\eqref{eq:eq9}, rather than attempting to minimize a combination of the two loss functions simultaneously. This selective approach ensures that the optimization process is more targeted and effective. The attack generation is in Algorithm~\ref{alg::conjugateGradient}. $\mu_{2}$ is a hyperparameter.

{\bf Attack Robustness Enhancement.} To enhance the attack robustness of \system, particularly in physical-world scenarios, we incorporate the Expectation over Transformation (EoT)~\cite{athalye2018synthesizing, eykholt2018robust, jia2022fooling, Wang_2023_ICCV} illustrated in Fig.~\ref{fig:fig_overview}. This approach involves applying various transformations, including position shifting, rotation, color modification, and so on. Additionally, to further enhance the attack robustness, we also provide a design option leveraging a dual-patch attack strategy with a monitor. The details are in Appendix~\ref{sec:duel_patch}.

\section{Evaluation}
\label{sec:eval}

We provide evaluations of \system in this section.

\begin{table*}[t]
\tabcolsep 0.05in
\renewcommand{\arraystretch}{0.85}
    \caption{The effectiveness of attacks on four object detection (OD) models, i.e., ApoD, Y3, Y5, and YX, with four MOT algorithms, i.e., ApoT, BoT-SORT, ByteTrack, and StrongSORT. The evaluation metrics include the attack success rate (ASR) and the average number of frames to execute an effective attack (Frame \#).}

    \centering
    \begin{tabular}{cccccccccccc}

    \toprule
    & & & \multicolumn{4}{c}{BDD dataset~\cite{yu2020bdd100k}} & \multicolumn{4}{c}{KITTI dataset~\cite{Geiger2013IJRR}} \\
    \cmidrule(lr){4-7}
    \cmidrule(lr){8-11}
     \multirow{-2.5}{*}{\shortstack{Attack \\ scenario}}& \multicolumn{2}{c}{OD\textbackslash{}MOT}
      & ApoT & BoT-SORT & ByteTrack & StrongSORT & ApoT & BoT-SORT & ByteTrack & StrongSORT & Average\\
    \midrule
    & & ASR & 100\% & 100\% & 100\% & 90\% & 90\% & 100\% &100\%& 90\% & 96.3\%\\
    & \multirow{-2}{*}{ApoD}& Frame \# & 3.1 & 2.8 & 2.6 & 2.5& 2.4 & 2.4 & 2.7 & 2.6 & 2.6\\
    \cmidrule(lr){2-12}
        & & ASR & 100\% & 100\% & 100\% & 100\% & 100\% & 100\% & 100\% & 100\% & 100\%\\
        & \multirow{-2}{*}{Y3}& Frame \# & 3.2 & 2.9 & 3.4 & 2.7& 2.5 & 2.6 & 3.1 & 2.4 & 2.9\\
    \cmidrule(lr){2-12}
        & & ASR & 100\% & 100\% & 100\% & 100\% & 100\% & 100\% & 100\% & 100\% & 100\%\\
    & \multirow{-2}{*}{Y5}  & Frame \# & 3.1 & 2.8 & 2.9 & 2.6 & 2.7 & 2.3 & 3.0 & 2.9 &2.8\\
    \cmidrule(lr){2-12}
        & & ASR & 100\% & 100\% & 100\% & 100\% & 100\% & 100\% & 100\% & 90\% & 98.8\%\\
     \multirow{-10}{*}{Move-in}& \multirow{-2}{*}{YX}& Frame \# & 3.8 & 3.0 & 3.1 & 2.7 & 2.6 & 2.9 & 3.5 & 2.3& 3.0\\
    \midrule
    & & ASR & 100\% & 100\% & 100\% & 100\%& 90\% & 100\% & 100\% & 100\% & 98.8\%\\
    &  \multirow{-2}{*}{ApoD} & Frame \# & 3.6& 2.6 & 2.5 & 2.4 & 2.9 & 2.6 & 2.4 & 2.8 & 2.7\\
    \cmidrule(lr){2-12}
        & & ASR & 100\% & 100\% & 100\% & 100\% & 100\% & 100\% & 100\% & 100\% & 100\%\\
    & \multirow{-2}{*}{Y3}  & Frame \# & 4.0 & 2.8 & 2.7 & 2.2 & 2.9 & 2.4 & 2.4 & 2.2 &2.7\\
    \cmidrule(lr){2-12}
        & & ASR & 100\% & 100\% & 100\% & 100\% & 100\% & 100\% & 100\% & 100\% & 100\%\\
    &\multirow{-2}{*}{Y5} & Frame \# & 3.5 & 2.7 & 2.6 & 2.3 & 2.8 & 2.2 & 2.4 & 2.2 & 2.6\\
    \cmidrule(lr){2-12}
        & & ASR & 90\% & 90\% & 90\% & 90\% & 80\% & 90\% & 100\% & 100\%& 91.3\%\\
    \multirow{-10}{*}{Move-out}& \multirow{-2}{*}{YX}& Frame \# & 4.5 & 2.9 & 2.9 & 2.8 & 2.7 & 2.4 & 2.9 & 2.4 & 2.9\\
    \midrule
    \midrule
    & & ASR & 98.8\% & 98.8\% & 98.8\% & 97.5\% & 95.0\% & 98.8\% & 100\% & 97.5\% & 98.1\%\\
    \multicolumn{2}{c}{\multirow{-2}{*}{Average}} & Frame \# & 3.6 & 2.8 & 2.8 & 2.5 & 2.7 & 2.5 & 2.8 & 2.5 & 2.8\\
    \bottomrule
         
    \end{tabular}

    \label{tab:main_res}
    \vspace{-0.2cm}
\end{table*}

\begin{figure*}[t]
\centering
\hspace{0.1cm}
{\includegraphics[width=6in]{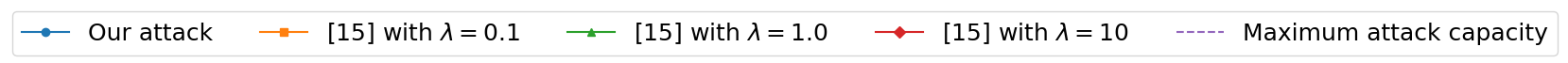}%
\label{fig_legend}}

\vspace{-0.4cm}
\subfloat[Y3, cov=0.1]
{\includegraphics[width=1.57in]{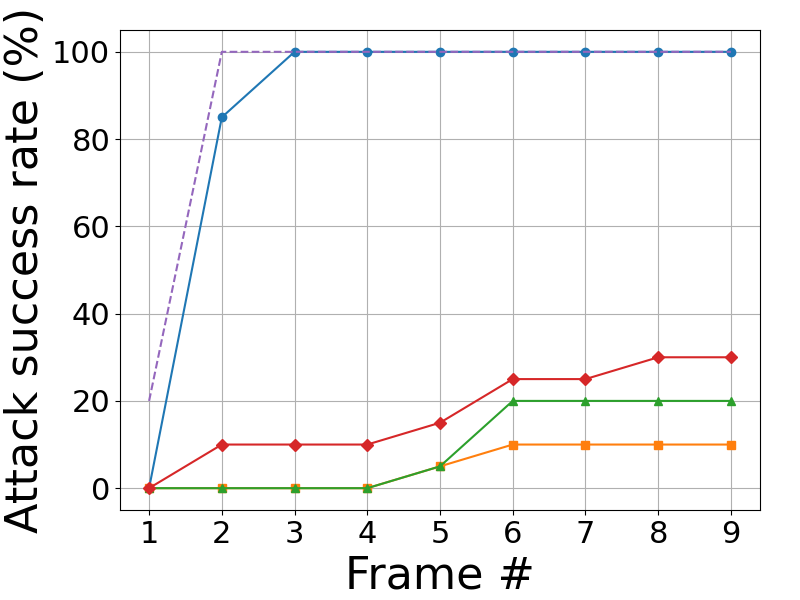}%
\label{fig_3a}}
\subfloat[Y5, cov=0.1]
{\includegraphics[width=1.57in]{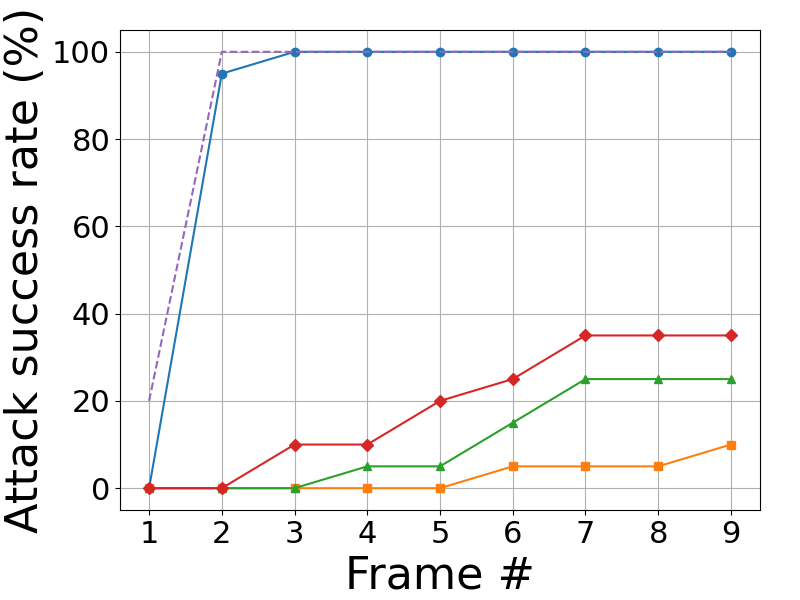}%
\label{fig_3b}}
\subfloat[ApoD, cov=0.1]
{\includegraphics[width=1.57in]{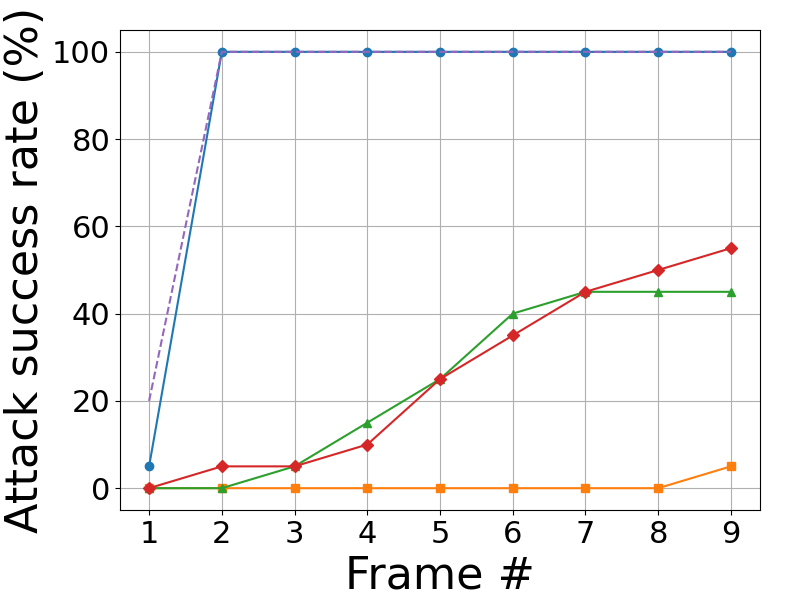}%
\label{fig_3c}}
\subfloat[YX, cov=0.1]
{\includegraphics[width=1.57in]{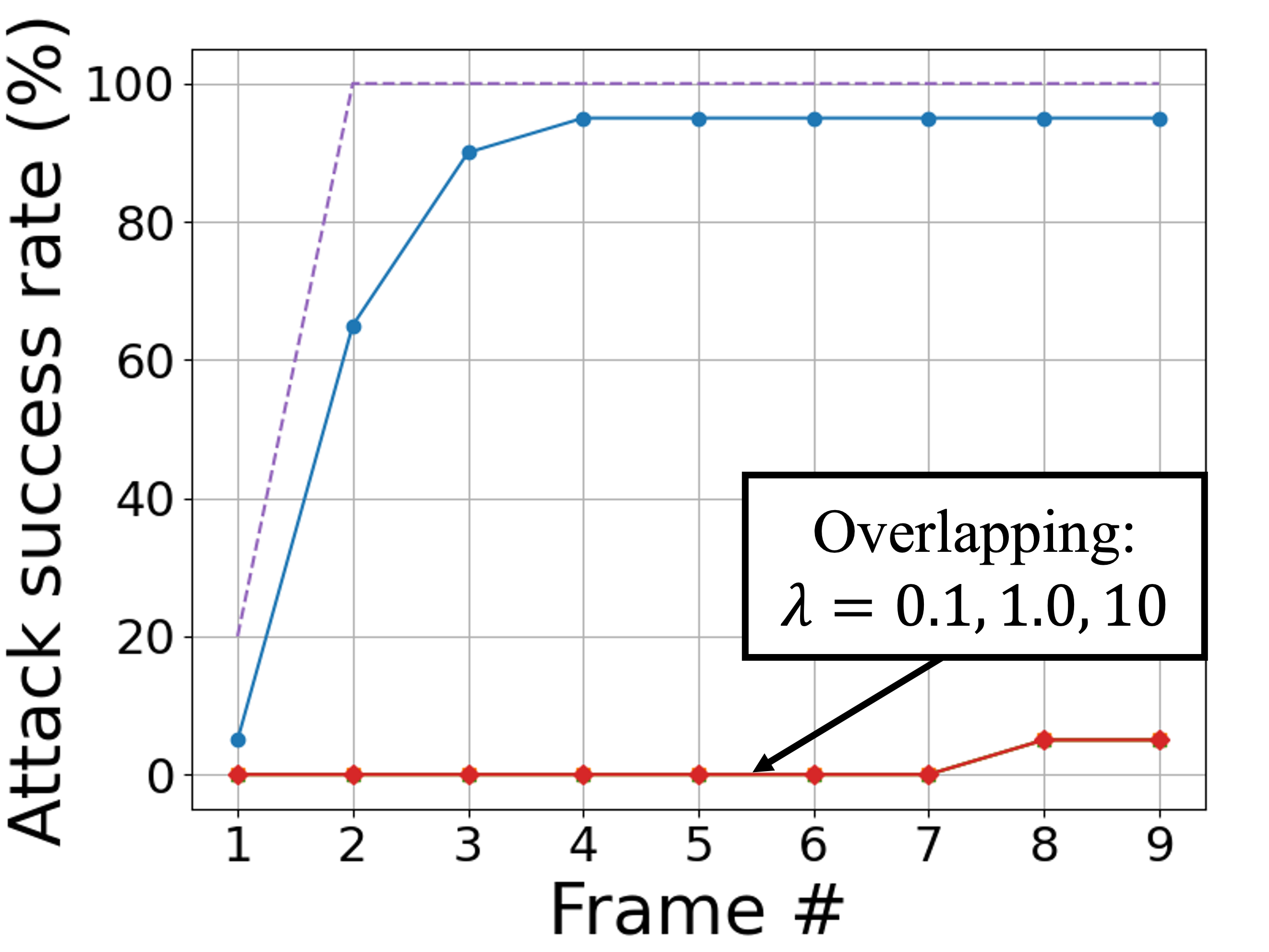}%
\label{fig_3d}}

\vspace{-0.4cm}
\subfloat[Y3, cov=1]
{\includegraphics[width=1.57in]{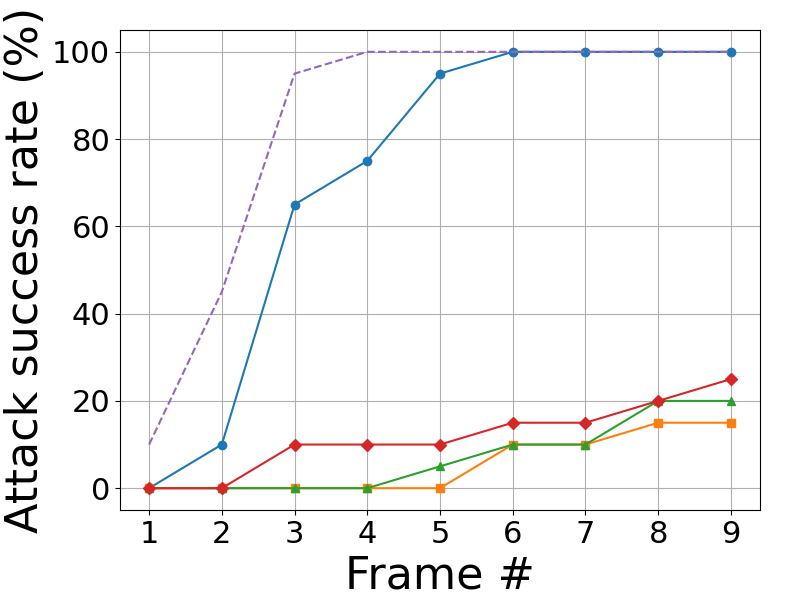}%
\label{fig_3e}}
\subfloat[Y5, cov=1]
{\includegraphics[width=1.57in]{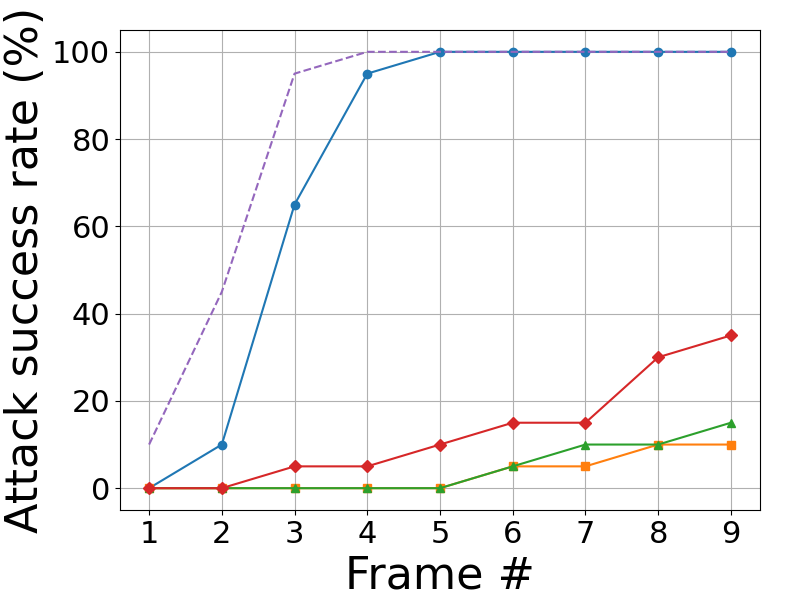}%
\label{fig_3f}}
\subfloat[ApoD, cov=1]
{\includegraphics[width=1.57in]{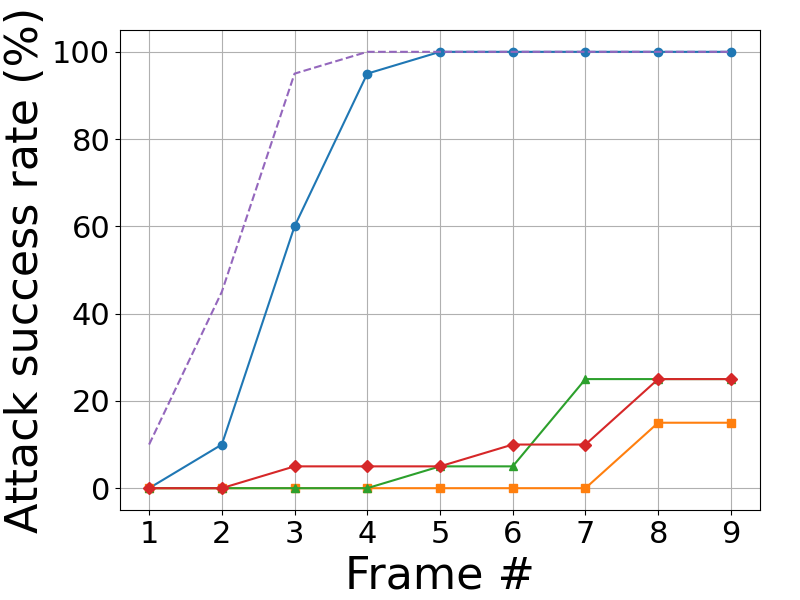}%
\label{fig_3g}}
\subfloat[YX, cov=1]
{\includegraphics[width=1.57in]{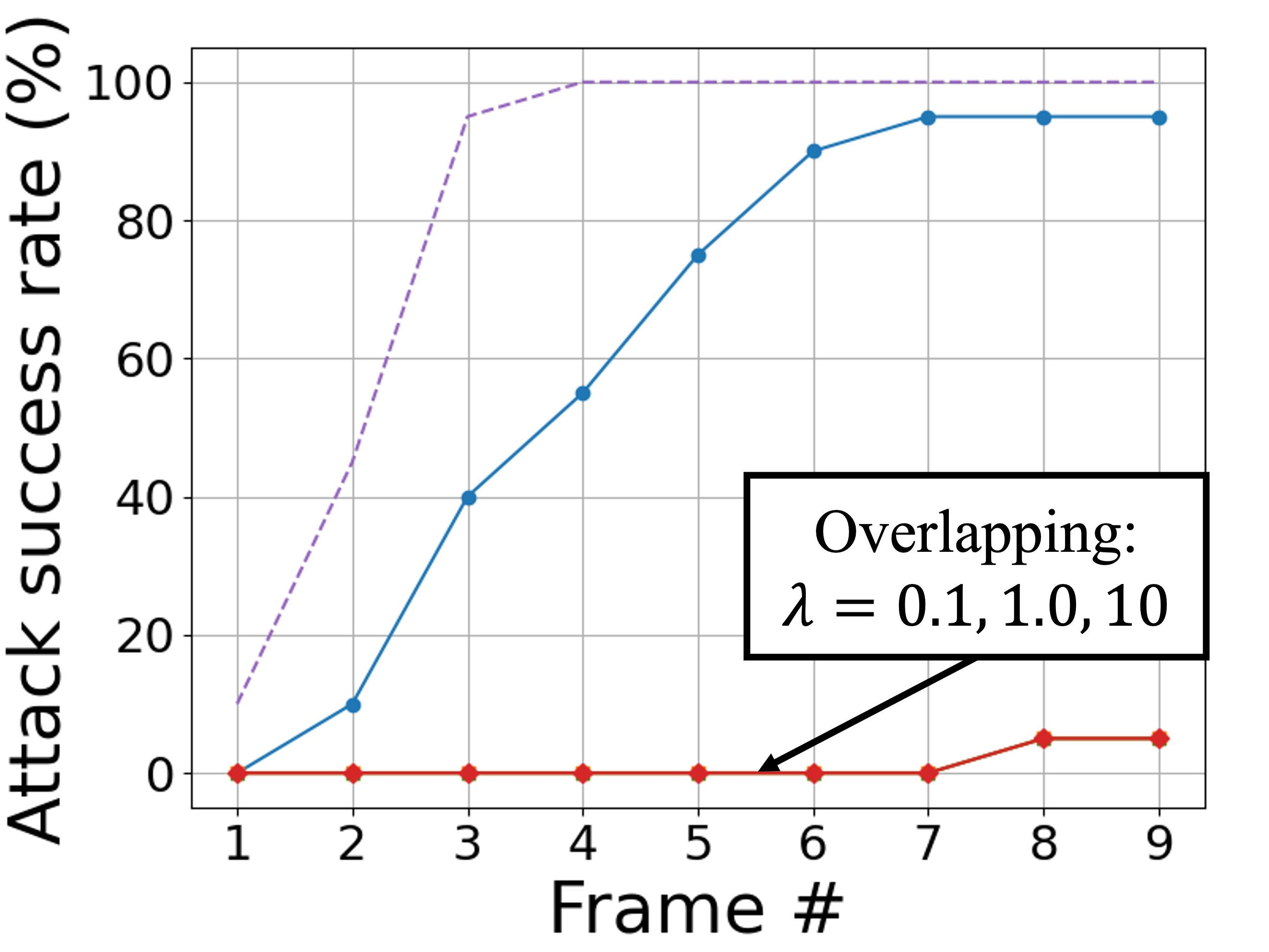}%
\label{fig_3h}}

\vspace{-0.4cm}
\subfloat[Y3, cov=10]
{\includegraphics[width=1.57in]{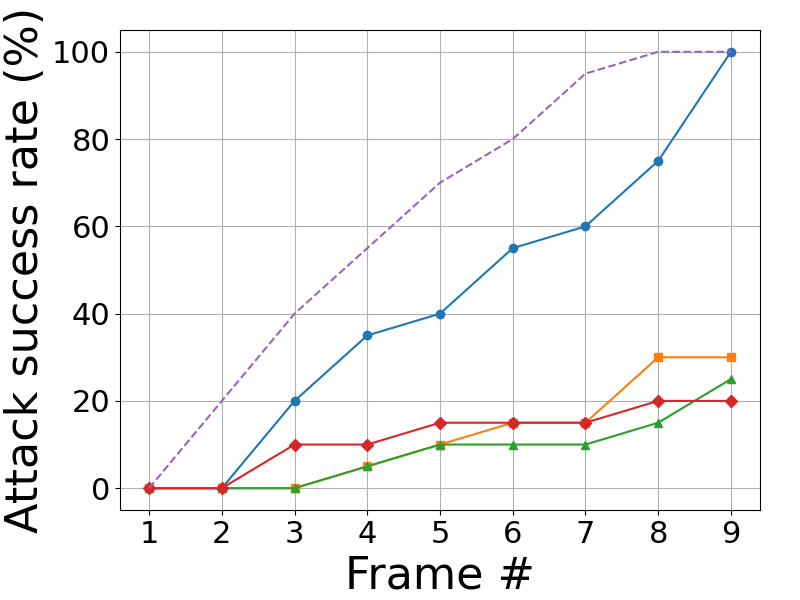}%
\label{fig_3i}}
\subfloat[Y5, cov=10]
{\includegraphics[width=1.57in]{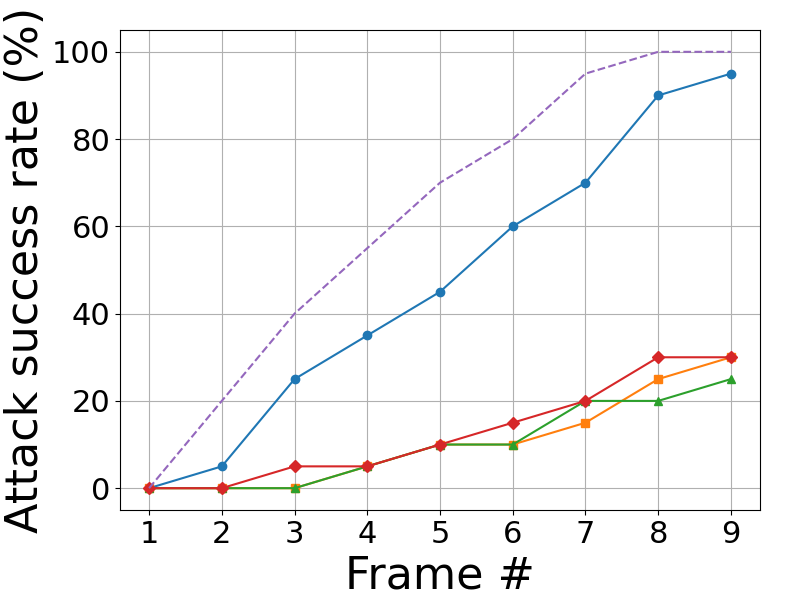}%
\label{fig_3j}}
\subfloat[ApoD, cov=10]
{\includegraphics[width=1.57in]{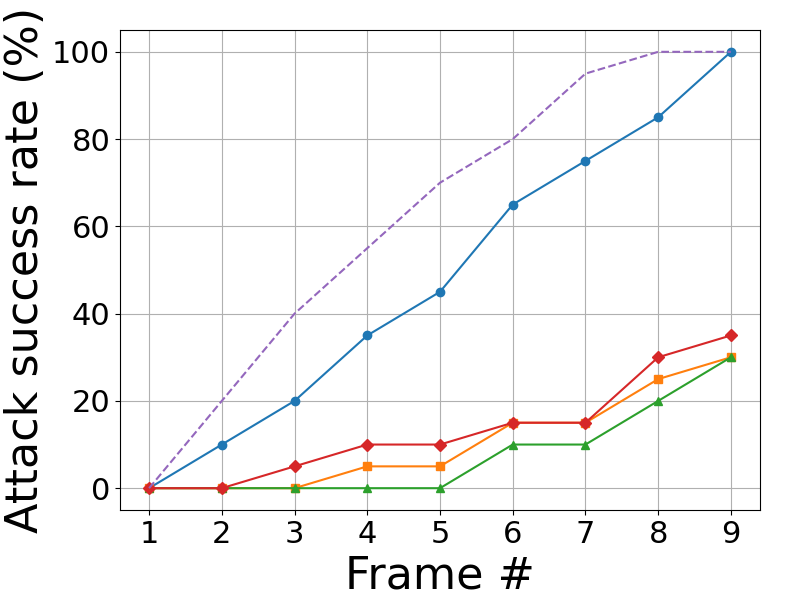}%
\label{fig_3k}}
\subfloat[YX, cov=10]
{\includegraphics[width=1.57in]{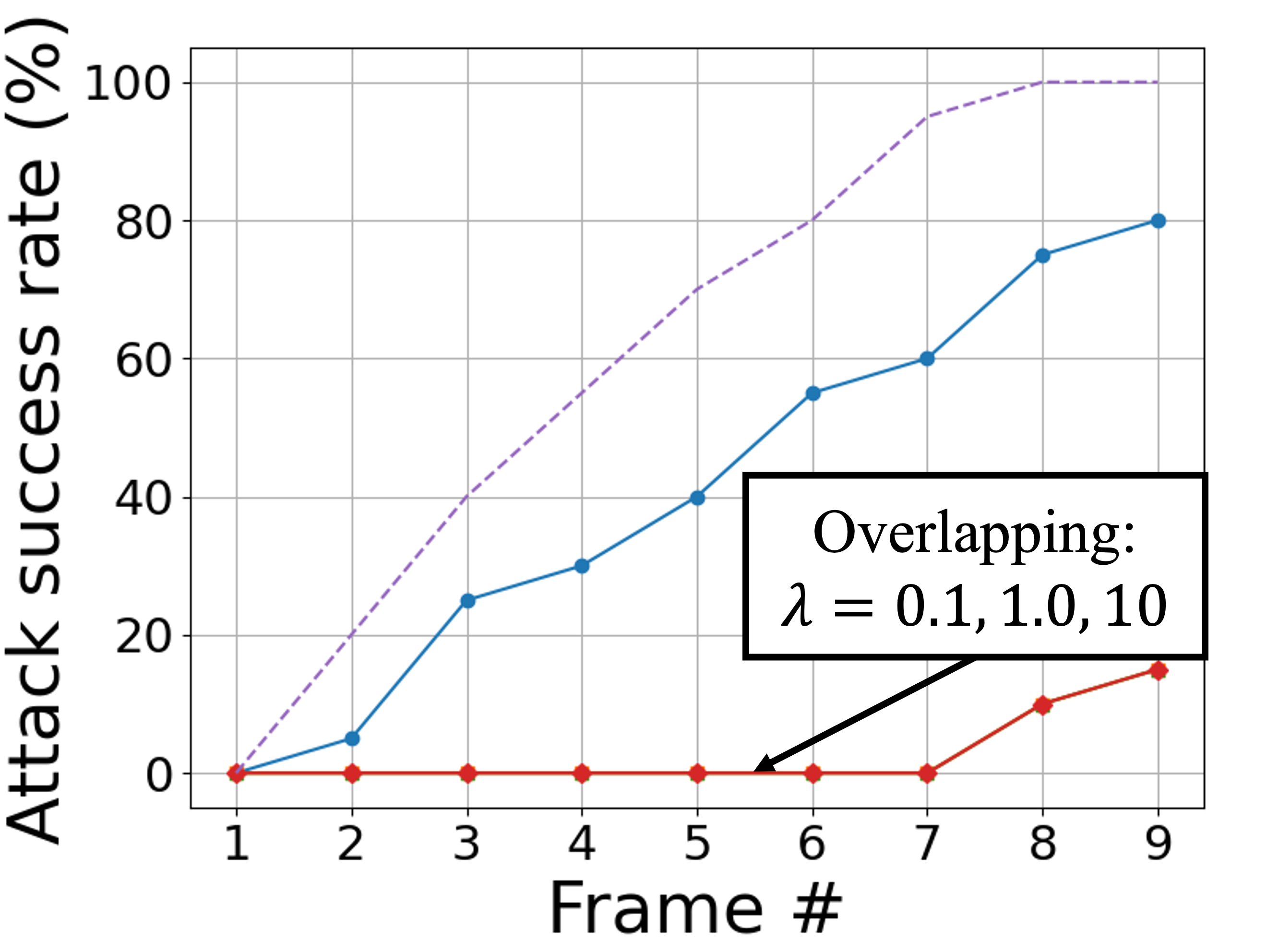}%
\label{fig_3l}}

\caption{Comparison between our attack and the baseline attack~\cite{jia2020fooling} under four different object detection models (Y3, Y5, ApoD, and YX) with three different parameter values of ApoT (cov = 0.1, 1, 10). $\lambda$ is the hyperparameter in~\cite{jia2020fooling}. Maximum attack capability assumes the attacker can arbitrarily control BBOX locations.}
\label{fig:attack}
\vspace{-0.2cm}
\end{figure*}

\begin{figure*}[!t]
\centering
\includegraphics[width=0.95\linewidth]{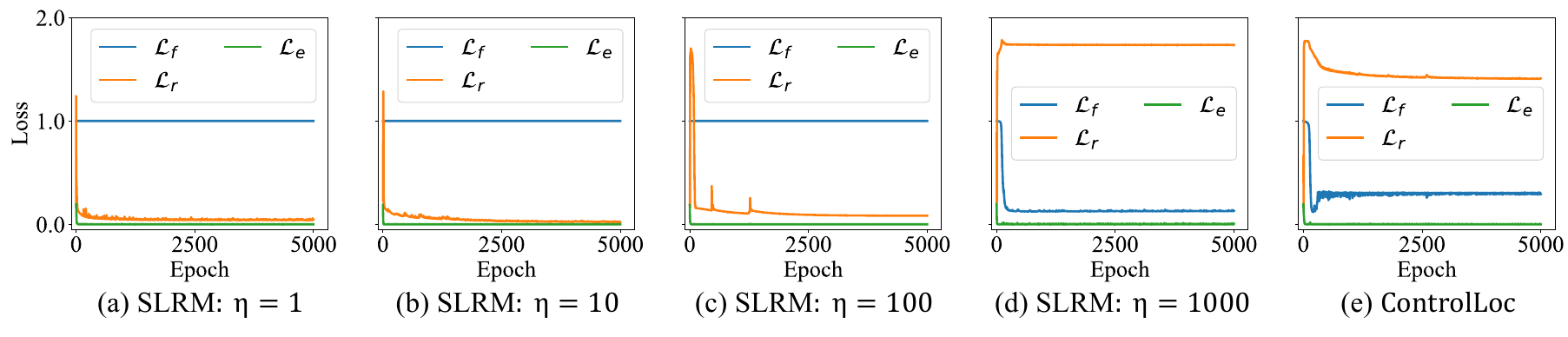}
\vspace{-0.3cm}
\caption{Comparison of loss value between our optimization method in \system and the standard Lagrangian relaxation method (SLRM) with different hyperparameter values $\eta$. The detailed loss designs are in~\S\ref{sec:loss_design}: Equation~\eqref{eq:score_loss} and Equation~\eqref{eq:regr_loss}.}
\label{fig:optimization}
\vspace{-0.2cm}
\end{figure*}

\subsection{Evaluation Methodology and Setup}
\label{sec:eval_method_setup}

{\bf AD Perception.} We include different AD perception systems, i.e., different object detection models and MOT. For object detection, we encompass both anchor-based and anchor-free detectors. Our examination mostly leverages algorithms in open-source industry-grade full-stack AD systems to affirm the practicality and representativeness of our findings. We select a variety of object detection models, including the Baidu Apollo Object Detection (ApoD)~\cite{apollo}; YOLO v3 (Y3) ~\cite{redmon2018yolov3} as incorporated in Autoware.AI~\cite{kato2018autoware}; YOLO v5 (Y5)~\cite{Jocher2022yolov5} which is highlighted in recent security research on AD~\cite{jia2022fooling}; and YOLOX (YX)~\cite{ge2021yolox}, an anchor-free detector in the latest Baidu Apollo Beta. 
For MOT, our focus extends to leading and representative algorithms that underscore the diversity and advancement in the field. 
This includes the Baidu Apollo MOT (ApoT)~\cite{apollo}; BoT-SORT~\cite{aharon2022bot}; ByteTrack~\cite{zhang2022bytetrack},
and StrongSORT~\cite{du2023strongsort}. We all use their default configurations. 

{\bf Datasets.} We select two widely recognized datasets in the AD research~\cite{cao2021invisible, jia2020fooling, yu2020bdd100k, Geiger2013IJRR}: the Berkeley Deep Drive (BDD) dataset~\cite{yu2020bdd100k} and the KITTI dataset~\cite{Geiger2013IJRR}. Within the BDD dataset, we randomly chose 20 clips specifically for their relevance to our attack goals: 10 clips are for the object move-in scenario, and another 10 are chosen for the object move-out scenario. A similar selection process is applied to the KITTI dataset, where we select 10 clips each for the move-in and move-out scenarios. We manually identify a target vehicle within each clip. To align our study with realistic conditions, we impose restrictions on the size of the adversarial patch, for which on average, our patch size is only 12\% of the target vehicle in pixels. After the double-blind review process, we plan to make the data publicly available to facilitate further research within this field.

\subsection{Attack Effectiveness}
\label{sec:eval_attack_effect}
{\bf Evaluation Metrics.} The success of the attack is defined as \textit{the attack is considered successful when, at the end of the attack, the detection BBOX of the target object can no longer be associated with any existing trackers}. Such metric is widely used in the security analysis of tracking~\cite{jia2020fooling, muller2022physical}. We measure the attack success rate (ASR) and the average number of frames to conduct an effective attack 
(Frame \#). 
Note that the Frame \# is within the attack successful cases.

{\bf Results.} The attack effectiveness on four object detectors and four MOT algorithms across two datasets, aiming for two specific attack goals, is detailed in Table~\ref{tab:main_res}. The attack boasts an average success rate of 98.1\% and necessitates an average of 2.8 frames to achieve efficacy in general. Among the MOT algorithms evaluated across the two datasets, ApoT emerges as the most robust one, evidenced by its lowest average attack success rate of 96.9\% and the highest average of 3.2 frames required for a successful attack. These findings suggest that attacking ApoT demands a higher frame count and has a lower attack success rate, rendering it less vulnerable compared to other MOT algorithms. Regarding object detection, YX demonstrates the lowest attack success rate at 95.1\% and requires the highest average of 3.0 frames for a successful attack. This robustness could be attributed to its anchor-free object detection design, which appears more robust against hijacking attacks. Within the anchor-based object detection models, ApoD shows the lowest attack success rate at 97.6\%, suggesting that the design of object detection and MOT in Apollo tends to be more robust. Note that YX is also adapted in Apollo as introduced in~\S~\ref{sec:eval_method_setup}. An additional observation is that the move-in attack achieves a higher success rate of 98.8\% but generally requires more frames (average of 2.8) compared to the move-out attack, which has a success rate of 97.5\% with an average of 2.7 frames. This suggests that, although move-in attacks might be easier to successful than move-out attacks, the latter tend to reach attack goals faster within successful cases. From Table~\ref{tab:main_res}, StrongSORT exhibits greater robustness compared to others, except ApoT. This is likely due to a Noise Scale Adaptive Kalman filter~\cite{du2023strongsort} design, which adjusts measurement noise covariance based on confidence scores of detection results. 

{\bf Takeaway:} \system can successfully achieve the attack goals of hijacking attacks with a 98.1\% average attack success rate across various AD perceptions and datasets.


\subsection{Comparison with Baselines}
\label{sec:baseline}

\subsubsection{Comparison with Prior Attack}
We compare our \system to a representative baseline attack~\cite{jia2020fooling}.

{\bf Methodology and Setup.} For the visual perception pipeline, we chose different object detectors coupled with ApoT due to their adoption in an industry-grade full-stack AD system. The evaluation utilizes the BDD dataset as outlined in~\S\ref{sec:eval_method_setup}. Following the methodology of prior research as our baseline~\cite{jia2020fooling}, we employ $\lambda$ to denote the weighting factor between two loss functions in the baseline method, $\mathcal{L}_1$ and $\mathcal{L}_2$, thus defining the combined loss function as $\mathcal{L} = \mathcal{L}_1 + \lambda \cdot \mathcal{L}_2$. The $\mathcal{L}_1$ is for erasure and $\mathcal{L}_2$ is for fabrication. We also explore the impact of varying $\lambda$ values, specifically 0.1, 1.0, and 10.0. Additionally, for ApoT, we investigate its performance across different tracking parameters, notably $\mathrm{cov}$: noise covariance in Kalman filter~\cite{jia2020fooling} following the same evaluation setup as the baseline attack~\cite{jia2020fooling}. 

{\bf Results.} The results, as depicted in Fig.~\ref{fig:attack}, unequivocally demonstrate the superior efficacy of \system, achieving an impressive 99.4\% attack success rate on Y3, ApoD, and Y5 models, and a 90\% attack success rate on the YX model. This starkly contrasts with the outcomes from existing research~\cite{jia2020fooling}, which has an 8.3\% attack success rate on the YX model and 24.8\% on the other models tested. This substantial discrepancy underscores the enhanced capability of our \system to manipulate the target object's position effectively, thereby hijacking its tracker. A critical observation from our analysis reveals that prior research~\cite{jia2020fooling} tends to fail in maintaining the target's BBOX: at low $\lambda$ values, leading to its disappearance, or conversely, at high $\lambda$ values, resulting in no significant change or generating multiple BBOXes. In stark contrast, our \system demonstrates remarkable effectiveness and robustness to different $cov$ values of ApoT. In certain instances, \system achieves similar performance to maximum attack capacity, which assumes the attacker can arbitrarily manipulate the BBOXes. 

{\bf Takeaway:} \system outperforms hijacking attack baseline~\cite{jia2020fooling} on AD perception by fourfold improvements.

\subsubsection{Comparison with Traditional Optimization} 
This part compares our novel optimization method with the traditional standard Lagrangian relaxation method (SLRM) in this hijacking attack context. Our method, delineated in Equation~\eqref{eq:eq9}, diverges from SLRM, which merges score loss ($\mathcal{L}_s$) and regression loss ($\mathcal{L}_r$) using a hyperparameter $\eta$ in the form $\mathcal{L}_r + \eta \cdot \mathcal{L}_s$. Notably, the score loss encompasses two components, $\mathcal{L}_f$ and $\mathcal{L}_e$, as specified in Equation~\eqref{eq:score_loss}. To facilitate a detailed comparison, we use $\mathcal{L}_f$ and $\mathcal{L}_e$ for $\mathcal{L}_s$. Previous research leveraging SLRM~\cite{jia2020fooling} demonstrates its inadequacy in generating effective adversarial patches for tracker hijacking. This limitation is illustrated through the three losses, $\mathcal{L}_r$, $\mathcal{L}_f$, and $\mathcal{L}_e$, which fail to optimize simultaneously under varying hyperparameter $\eta$ settings, as depicted from Fig.~\ref{fig:optimization} (a) to (d). 
The primary challenge arises from the low initial score of the fabricated BBOX ($B_f$), resulting in a correspondingly weak gradient. Thus, SLRM hinders the minimization of $\mathcal{L}_f$, particularly when with high regression loss. 
This typically leads to a negligible reduction in $\mathcal{L}_f$, as evidenced in Fig.~\ref{fig:optimization} (a) to (c), where $\mathcal{L}_f$ barely decreases unless $\eta$ is substantially increased, for example, to around 1000, as shown in Fig.~\ref{fig:optimization} (d). However, elevating the $\eta$ introduces a new problem: the regression loss ($\mathcal{L}_r$) fails to be well optimized, shown in Fig.~\ref{fig:optimization} (d). This damages the attack's effectiveness, preventing the fabricated BBOX from associating with the target tracker. However, our optimization approach successfully mitigates these issues, showing its efficacy in Fig.~\ref{fig:optimization} (e), satisfying condition in Equation~\eqref{eq:eq3}.

{\bf Takeway:} Our newly proposed optimization approach can simultaneously optimize multiple loss functions, showing a better fit for hijacking attack scenarios than traditional standard Lagrangian relaxation methods.

\subsubsection{Baseline Evaluation for Stage I in~\S\ref{sec:patch_location}.} This part assesses the attack effectiveness benefit from Stage I by comparing two scenarios: utilizing Stage I for patch location preselection and employing a random patch location on the rear of the vehicle. To perform a fair comparison, we conduct 1,000 iterations for each attack generation. Specifically, for the Stage I scenario, we involve 20 iterations to determine the optimal patch location. The results, in Table~\ref{tab:ablation}, reveal that attacks employing Stage I achieve an average attack success rate of 85.0\% across two attack goals, whereas those with a random patch location exhibit a significantly lower attack success rate of 20.0\%. This, thus, underscores the importance of Stage I in \system. 

{\bf Takeway:} Stage I in \system can significantly improve the attack effectiveness compared to the random patch location baseline by more than fourfold improvements.

\begin{table}[t]
\renewcommand{\arraystretch}{0.85}
    \caption{ASR comparison between our patch location preselection (\S\ref{sec:patch_location}) in \system and a random location preselection on the rear of the vehicle.}
    \centering
    \begin{tabular}{ccccc}

    \toprule
    & \multicolumn{2}{c}{Random} & \multicolumn{2}{c}{Ours} \\
    \cmidrule(lr){2-3}
    \cmidrule(lr){4-5}
    & Move-in & Move-out& Move-in & Move-out \\
    \midrule
    ASR & 20\% & 20\% & \textbf{90\%} & \textbf{80\%} \\

    \bottomrule
         
    \end{tabular}

    \label{tab:ablation}
    \vspace{-0.4cm}
\end{table}

\begin{table}[t]
\tabcolsep 0.1in
\renewcommand{\arraystretch}{0.85}
    \caption{Physical-world attack evaluation regarding ASR under different outdoor lighting conditions/backgrounds with two hijacking directions: from left to right and from right to left. The results are averaged over 5 videos.}
    \centering
    \begin{tabular}{ccccc}

    \toprule
    Attack& \multicolumn{2}{c}{From left to right} & \multicolumn{2}{c}{From right to left}\\
    \cmidrule(lr){2-3}
    \cmidrule(lr){4-5}
     scenario& Benign & \system & Benign & \system \\
    \midrule
    \multicolumn{5}{c}{\textbf{Inside the parking structure ($\sim$600 lux)}}\\
    \midrule
    Move-in & 0\% & \textbf{80\%} & 0\% & \textbf{80\%} \\
    Move-out & 0\% & \textbf{60\% }& 0\% & \textbf{60\%} \\
    \midrule
    \multicolumn{5}{c}{\textbf{On the rooftop of parking structure ($\sim$30,000 lux)}}\\
    \midrule
    Move-in & 0\% & \textbf{100\%} & 0\% & \textbf{80\%} \\
    Move-out & 0\% & \textbf{80\%} & 0\% & \textbf{80\%} \\

    \bottomrule
         
    \end{tabular}

    \label{tab:physical}
    \vspace{-0.2cm}
\end{table}

\subsection{Physical-World Attack Evaluation
}
\label{sec:physical}
While the results in previous sections show the effectiveness of \system in the digital space, 
it is unclear whether the effectiveness can be maintained once the adversarial patch is printed and deployed in the real world. Thus, we further assess the physical-world attack realizability. 

\begin{figure}[!t]
\centering
\includegraphics[width=\linewidth]{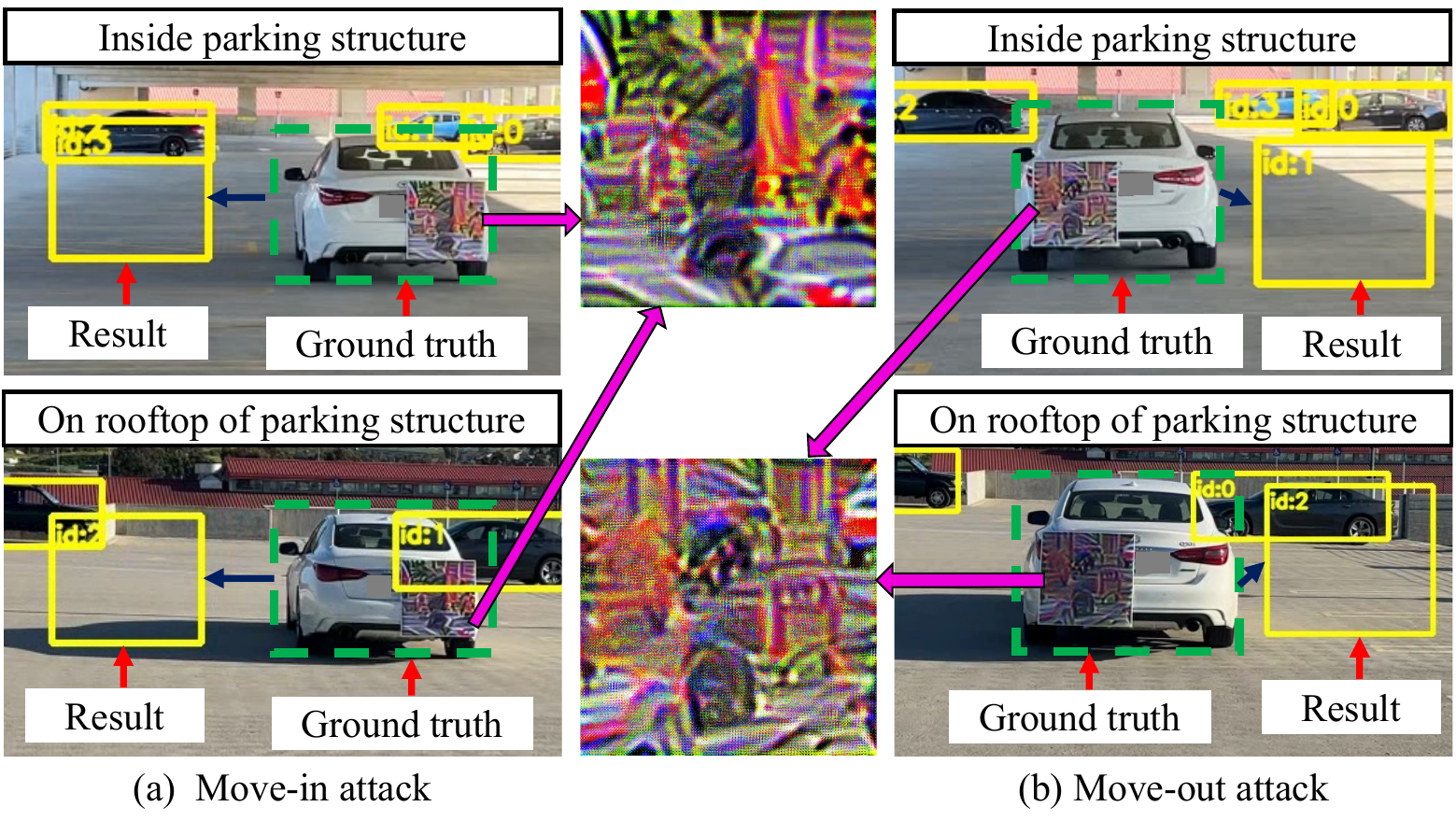}
\vspace{-0.5cm}
\caption{The visualizations of physical-world attacks with two adversarial patches for (a) move-in, from right to left, and (b) move-out, from left to right, with different backgrounds and lighting conditions.}
\label{fig:patch_vis}
\vspace{-0.4cm}
\end{figure}

{\bf Evaluation Setup and Methodology.} In physical-world attack evaluation, we explore the hijacking attack from both lateral directions: left-to-right and right-to-left. For each direction, we aim to achieve both move-in and move-out attack goals depending on the positioning of the vehicle. Our evaluation encompasses all possible scenarios. We utilize an adversarial patch 29 inches by 29 inches, which is smaller than the physical monitor used in existing research on AD security~\cite{man2023person}. This patch is placed on the rear of a standard sedan. Note that the patch size and location can be controlled by the attacker, as the patch is placed on the attacker’s car. The placement of the patch is strategically determined by our patch location preselection methodology in~\S\ref{sec:patch_location}. Videos are captured under clear and sunny weather conditions during the daytime, encompassing both benign and adversarial scenarios. The camera used for these recordings has the same configurations—such as focal length and video resolution—as the camera used in Baidu Apollo~\cite{apollo}. For each scenario, we collect five video clips for analysis. The experiments are conducted in two distinct outdoor light environments within our institute: the interior of a parking structure (approximately 600 lux) and the rooftop of the parking structure (approximately 30,000 lux), to assess the impact of varying outdoor light conditions. For object detection and MOT, we utilize the systems implemented in Baidu Apollo, specifically ApoD and ApoT, due to their representativeness. Two unique adversarial patches are printed, one for each direction of hijacking, to test their effectiveness under different outdoor light conditions/backgrounds and achieve different attack goals based on the vehicle location.

{\bf Results and Visualization.} The effectiveness of \system, under variations in outdoor light conditions, background, hijacking directions, and attack goals, is presented in Table~\ref{tab:physical}. Our \system achieves a 77.5\% average attack success rate, while in all benign scenarios, the perception functions normally, evidenced by a 0\% attack success rate. Notably, brighter outdoor light conditions have better effectiveness, yielding an 85.0\% attack success rate compared to a 70.0\% attack success rate in darker conditions. This suggests that our attack is more potent during daytime, which is a common time period. As for the attack goals, the move-in attack achieves a higher success rate of 80.0\%, making it more effective than the move-out attack, which has a success rate of 75.0\%. This observation is consistent with the findings from digital-space evaluations, as detailed in~\S\ref{sec:eval_attack_effect}. Additionally, we observe that attacks hijacking direction from left to right exhibit a higher attack success rate (80.0\%) than those from right to left (75.0\%). This asymmetry may stem from the slight discrepancies in patch placement within the physical environment, given the inherent challenges of precisely controlling the patch location. Nevertheless, with a lower bound average attack success rate of 75.0\%, our findings indicate that the patch placement method can accommodate minor errors.

Fig.~\ref{fig:patch_vis} provides visualizations of our attack evaluations in the physical world with two patches, including two hijacking directions, two attack goals, and different outdoor light conditions and backgrounds. Fig.~\ref{fig:patch_vis} (a) and (b) depict the move-in and move-out attacks, respectively. Intriguingly, our Stage I methodology tends to place the patch on the side opposite to the direction of the hijacking. This placement, determined by the Stage I methodology, reveals a notable characteristic of the hijacking attack on AD perception.

{\bf Takeaway:} \system can successfully achieve the hijacking attack in the real world with a 77.5\% average attack success rate across different outdoor light conditions, backgrounds, hijacking directions, and attack goals.

\subsection{System-Level Attack Effect Evaluation}
\label{sec:end-to-end}
To understand the safety consequences of \system, we evaluate it on concrete scenarios in an AD simulator.

{\bf Evaluation Setup and Methodology.} To study the AD system-level attack effects of \system, we perform an attack evaluation on Baidu Apollo~\cite{apollo}, an industry-grade full-stack AD system, using LGSVL simulator~\cite{rong2020lgsvl}, a production-grade AD simulator. Our experiments are conducted on the Borregas Ave map and the Lincoln2017MKZ AD vehicle. To enhance the perception fidelity of simulators, we model the location of the tracker after it has been hijacked and inject it into the AD system from our physical-world attack evaluation results in~\S\ref{sec:physical} including the two different light conditions and background. Our evaluation focuses on two representative scenarios as shown in Fig.~\ref{fig:ill}. In Fig.~\ref{fig:ill}, the blue vehicle is the victim AD vehicle and the blue and gray lines are the trajectories of the two vehicles. 
S1 (Fig.~\ref{fig:ill} (a)) is a common scenario for other vehicles to park on the side of the road and S2 (Fig.~\ref{fig:ill} (b)) is another common driving scenario. We perform 20 runs on each scenario with different speeds: 20 km/h and 40 km/h.

{\bf Results.} The outcomes are summarized in Table~\ref{tab:simulation}. Our \system achieves an average AD system-level attack effectiveness rate of 81.3\% for critical scenarios such as vehicle collisions or unnecessary emergency stops while maintaining normal operation in benign situations with a 0\% incidence of attack effects on the AD system. The efficacy of the move-in attack (S1) at 91.3\% is notably superior to that of the move-out attack (S2), which has a 71.3\% rate. Notably, the attack effectiveness at high speeds (40 km/h) reaching 77.5\% surpasses that at lower speeds (20 km/h), which is 65.0\% in the move-out scenarios. This is critical as high-speed scenarios pose significant safety risks.

{\bf Takeaway:} \system can induce AD system-level impacts, like collisions and unnecessary emergency stops, with 81.3\% average effectiveness in a production AD simulator.


\begin{table}[!t]
\tabcolsep 0.05in
\renewcommand{\arraystretch}{0.85}
    \caption{AD system-level evaluation (vehicle collision or unnecessary emergency stop rate) under different speeds using Baidu Apollo and LGSVL. 20 runs for each cell.}
    \centering
    \begin{tabular}{ccccccccc}

    \toprule
    & \multicolumn{4}{c}{Move-in} & \multicolumn{4}{c}{Move-out}\\
    \cmidrule(lr){2-5}
    \cmidrule(lr){6-9}
    & \multicolumn{2}{c}{600 lux} & \multicolumn{2}{c}{30,000 lux} & \multicolumn{2}{c}{600 lux} & \multicolumn{2}{c}{30,000 lux}\\
    \cmidrule(lr){2-3}
    \cmidrule(lr){4-5}
    \cmidrule(lr){6-7}
    \cmidrule(lr){8-9}
    Speed (km/h) & 20  & 40  & 20  & 40  & 20  & 40  & 20  & 40 \\
    
    \midrule
    Benign & 0\% & 0\% & 0\% & 0\% & 0\% & 0\% & 0\% & 0\% \\
    \textbf{\system} & 100\% & 90\% & 95\% & 80\% & 70\% & 80\% & 60\% & 75\%\\

    \bottomrule
         
    \end{tabular}
    \vspace{-0.2cm}

    \label{tab:simulation}
\end{table}

\begin{figure*}[!ht]
\centering
\includegraphics[width=\linewidth]{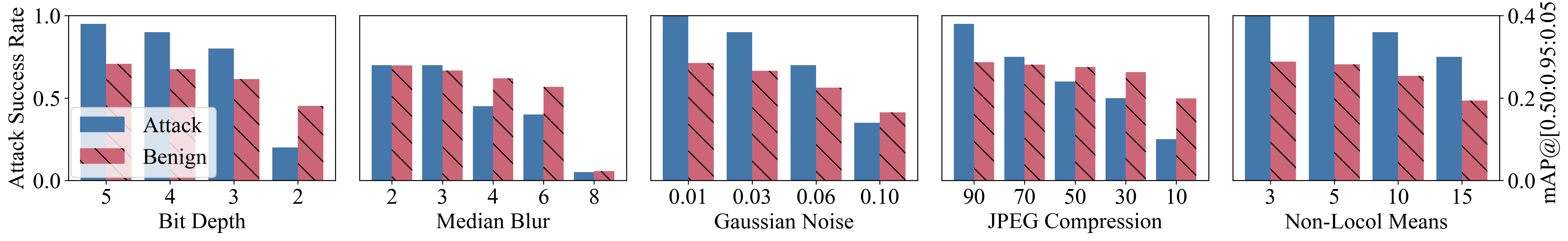}
\vspace{-0.6cm}
\caption{Attack effectiveness regarding attack success rate and model utility regarding mAP (mean Average Precision) of five common input transformation-based defenses. The x-axis represents the strength of each defense.}
\label{fig:defense1}
\vspace{-0.4cm}
\end{figure*}

\section{Discussion}
\label{sec:discussion}

\subsection{Defenses}
\label{sec:defense}

{\bf DNN-Based Defense.} 
Prior research has focused on enhancing the robustness of DNNs against adversarial attacks, aiming either to detect or mitigate these threats. Such efforts fall into two broad categories: certified defenses~\cite{li2023sok, xiang2023objectseeker,xiang2021patchguard} and non-certified defenses~\cite{xu2017feature, dziugaite2016study, zhang2019defending}. Certified defenses offer provable guarantees of robustness but are generally time-intensive, rendering them impractical for real-time systems, like AD systems. Furthermore, there is a notable absence of certified defenses specifically designed to defend against attacks on the entire AD perception. Thus, we evaluate several non-certified defense strategies: input-transformation defenses, which are directly adaptable. These include JPEG compression~\cite{dziugaite2016study}, bit depth reduction~\cite{xu2017feature}, Gaussian noise~\cite{zhang2019defending}, median blur~\cite{xu2017feature}, and non-local means~\cite{xu2017feature, zhang2020interpretable}. Due to their easily adaptable nature, these methods have been assessed in recent security studies~\cite{cao2021invisible, sato2021dirty, zhu2023tpatch, zhang2020interpretable}. We use the BDD dataset and the perception module in Baidu Apollo, i.e., ApoD and ApoT.

The effectiveness of these defense measures is quantified by the attack success rate, while the impact on benign performance is assessed using the mean Average Precision (mAP). As shown in Fig.~\ref{fig:defense1}, we observe that median blur can partially mitigate the attacks, particularly with large kernel sizes (e.g., 8). However, it remains possible for the attack to succeed, and importantly, applying such a filter also harms the model utility, which inevitably causes serious consequences in safety-critical applications, such as AD~\cite{zhu2023tpatch}. Thus, these defenses are not practically applicable.

{\bf Sensor Fusion Based Defense.} At the AD system level, employing multi-sensor fusion (MSF) for improving perception robustness, such as integrating LiDAR technologies, represents a strategic defensive approach. However, incorporating additional sensors like LiDAR substantially raises system costs. Thus, many AD systems primarily utilize camera-based perception, such as Tesla~\cite{jing2021too} and OpenPilot~\cite{openpilot}. Additionally, relying solely on MSF may not adequately defend against \system. This vulnerability is attributed to the potential for attackers to simultaneously attack all perception sources~\cite{cao2021invisible}. Furthermore, recent research~\cite{cheng2023fusion} shows the feasibility of attacking the MSF-based perception by attacking only camera-based perception. While the MSF may compromise the attack, its defensive potential remains to be systematically explored in future research.

{\bf Collision Avoidance System Based Defense.} Collision avoidance systems (CAS), like Autonomous Emergency Braking (AEB), use RADAR or ultrasonic sensors to prevent or reduce the severity of collisions~\cite{cao2021invisible, abdulhamid2020collision, kwag2006collision}. While helpful, they cannot fully prevent collisions or eliminate the need for robust defense methodologies against \system. First, AD systems must independently handle as many safety hazards as possible, rather than relying solely on CAS, which should serve as backup safety measures for emergencies. Second, CAS may have limited effectiveness against move-out attacks and are insufficient for defending against move-in attacks, making them inadequate against \system. Additionally, these systems are not perfect and can exhibit high false-negative rates, achieving only a 27\% reduction in bodily injury claim frequency and a 19\% reduction in property damage frequency~\cite{collisionavoidance, aeb}. Third, even if the system on the victim vehicle performs an emergency stop, it may not prevent rear-end collisions by following vehicles. Thus, while CAS provides a safety measure, they are not sufficient for complete or perfect collision avoidance. AD systems must be designed to handle safety hazards independently, and robust defenses are necessary.

\subsection{Ethics}
\label{sec:ethics}
When addressing the ethics related to the evaluation of physical-world attacks, it is imperative to underscore the evaluation taken to ensure both safety and responsibility. Our experimental setup is situated within a parking area in our institute, chosen for its controlled environment, which allows us to conduct tests in isolation, ensuring no unintended exposure to bystanders or other public roads. This can effectively avoid the risk of unintended consequences to the uninvolved public. Additionally, we confirm that no harm is caused to the commercial vehicles in our physical-world experiments. These vehicles are for data collection.

\subsection{Limitation and Future Work}
\label{sec:limit}
Despite promising outcomes in physical-world tests, the full impact of end-to-end attacks on AD systems, especially commercial vehicles like Tesla, remains to be fully understood. Constraints such as cost and safety lead us to use simulations~\cite{cao2021invisible}, a common industry practice, for preliminary AD system-level assessments~\cite{cao2021invisible, Wang_2023_ICCV}. More comprehensive and realistic evaluations can be a future work. The stealthiness of \system has not been thoroughly analyzed. Future efforts should focus on enhancing its inconspicuousness, potentially through advanced techniques such as content loss~\cite{zhu2023tpatch} or diffusion-based approaches~\cite{xue2024diffusion}. Our research predominantly examines one-stage detectors used in industry-grade AD systems. However, considering the existence and application of two-stage detection, systematic investigation can be a future direction for attack generality.

\section{Conclusion}
\label{sec:conclusion}

In this paper, we present \system, a novel physical-world adversarial patch attack to exploit vulnerabilities in AD perception including object detection and MOT. With a two-stage attack methodology, \system significantly outperforms the existing attack, achieving an impressive average success rate of 98.1\% across diverse AD perception systems. The effectiveness of \system is validated in real-world conditions, including real vehicle tests under different outdoor light conditions and backgrounds with a 77.5\% average attack success rate. AD system-level impact such as vehicle collisions and unnecessary emergency stops is evaluated using a production AD simulator.



%
\IEEEpeerreviewmaketitle






%


\bibliographystyle{IEEEtran}
\bibliography{main.bib}

\appendices
\section{Location of BBOXes for Gride-Based Object Detector}
\label{sec:yolo5}

\begin{figure}[!th]
\centering
\includegraphics[width=2. in]{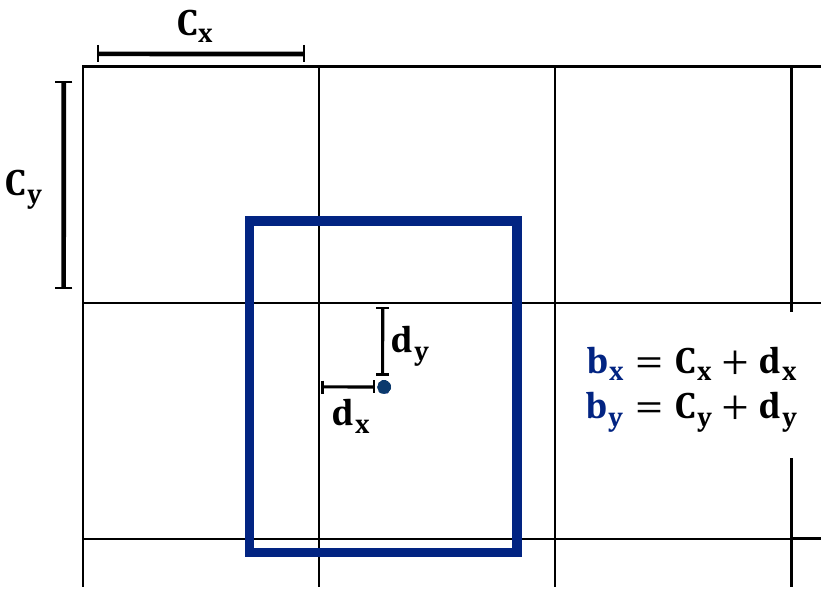}
\caption{The center coordinates of bounding boxes prediction method adopted by grid-based detection algorithms.}
\label{fig:fig_filter}
\vspace{-0.4cm}
\end{figure}

To ascertain the location of BBOXes for grid-based object detector, an offset is calculated from the top-left corner of each cell as shown in Fig.~\ref{fig:fig_filter}. For instance, YOLO v5~\cite{Jocher2022yolov5} determines the location coordinates relative to each cell's position. The detection model computes $t_x$ and $t_y$ for each BBOX within the output feature map. If a cell is positioned away from the image's top-left corner by ($c_x$, $c_y$), these predictions will be adjusted based on Equation~\eqref{eq:eq1}.
\begin{equation}
\label{eq:eq1}
\begin{array}{c}
b_x = d_x + c_x, \ \mathrm{where} \ \ d_x = \sigma(t_x)\\
b_y = d_y  + c_y, \ \mathrm{where} \ \ d_y = \sigma(t_y)
\end{array}
\end{equation}
where $\sigma(.)$ is the Sigmoid function. This process, utilized by grid-based detectors, calculates the offset $(d_x,d_y)$ to the center coordinates $(b_x,b_y)$ of each BBOX, ensuring that the center of any predicted BBOX remains within the confines of its cell.

\section{Dual Patch Attack Robustness Enhancement}
\label{sec:duel_patch}
To enhance the robustness of the attack, particularly in scenarios where the BBOX erasure is not completely effective in the first several frames of hijacking—potentially causing the hijacked tracker to match the BBOX of the original object and thereby failing the attack—we propose a dual-patch attack strategy. This approach consists of one patch designed for hijacking and a second for making the object disappear. The disappearing patch attack can draw upon methodologies similar to previous studies~\cite{jia2022fooling, Wang_2023_ICCV, zhao2019seeing} detailed as following.

Utilizing the BBOX filter described in~\S\ref{sec:box_filter}, we can efficiently identify and remove the BBOXes that should be excluded. We denote these BBOXes as the set ${B^\prime_e}$. To achieve the goal, we reduce the scores of all BBOXes in ${B^\prime_e}$, as formulated in Equation~\eqref{eq:rem}, aligning with the methodology used in previous research~\cite{zhao2019seeing, Wang_2023_ICCV, jia2022fooling, eykholt2018robust}.
\begin{equation}
\begin{aligned}
\label{eq:rem}
\mathcal{L}_\mathrm{e^\prime} ~& = {\frac{1}{|B^\prime_e|}}\sum\limits_{{c}\in{B^\prime_e}} c_{conf}^2
\end{aligned}
\end{equation}
where the $c_{conf}$ is detailed in Equation~\eqref{eq:cconf}.

The attacker can display these patches on a TV monitor, a similar threat model demonstrated by prior research~\cite{man2023person}, enhancing attack robustness in cases where hijacking might be challenging. However, as demonstrated in~\S\ref{sec:physical} that in real world, we successfully achieve two attack goals using a single patch, this dual-patch design is not always necessary but can be a design option for attack enhancement.

\begin{figure*}[!th]
\centering
\includegraphics[width=\linewidth]{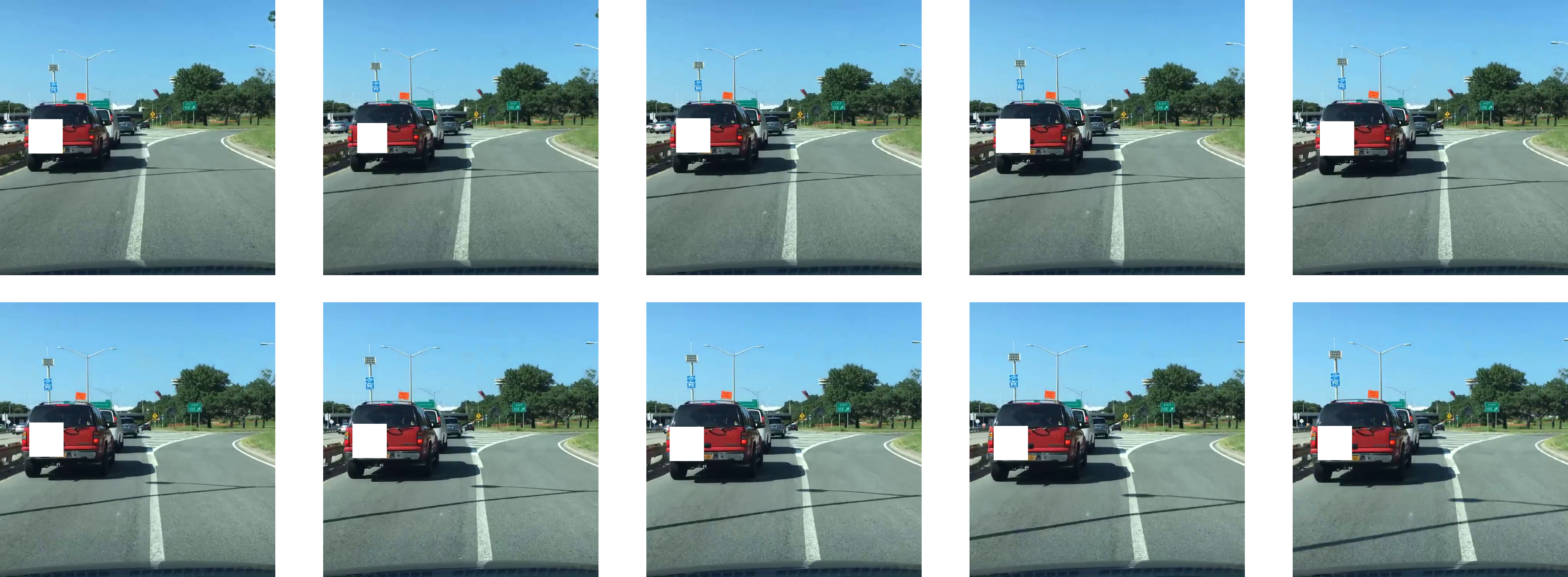}
\caption{Patch location under different frames. The white square in each image frame indicates the patch's optimal location.}
\label{fig:patch_loc}
\end{figure*}

\section{Patch Location under Different Frames}
We select several consecutive frames from a video clip and obtain their respective optimal patch positions via the method in Stage I, as illustrated in Fig.~\ref{fig:patch_loc}. The results demonstrate that the patch positions do not vary significantly across these frames and can be encompassed within the positional transformation distribution in the EoT. This observation indicates that our method is effective for continuous video frames.

\end{document}